\RequirePackage{fix-cm}
\documentclass[twocolumn]{svjour3}          % twocolumn
\smartqed  % flush right qed marks, e.g. at end of proof
\usepackage{graphicx}
\usepackage{multirow}
\usepackage{makecell}
\usepackage{subfigure}
\usepackage{booktabs} % for professional tables
\usepackage{amsmath,amssymb}
\usepackage{mathdots}
\usepackage{nicefrac}
\usepackage{xcolor}
\usepackage{pifont}
\usepackage{natbib}

\newcommand{\com}{\textcolor{black}} 
\newcommand{\comrevision}{\textcolor{black}} 
\newcommand{\comeng}{\textcolor{black}}

\begin{document} \sloppy

\title{\com{Discriminator} Feature-based Inference \\by Recycling the Discriminator of GANs}
% \subtitle{Do you have a subtitle?\\ If so, write it here}


\author{Duhyeon Bang$^{1,+}$  \and
        Seoungyoon Kang$^{1,+}$  \and
        Hyunjung Shim$^{1, *}$ 
}


\institute{Duhyeon Bang \at
              \email{duhyeonbang@yonsei.ac.kr}           %  \\
           \and
           Seoungyoon Kang \at
              \email{sy.kang@yonsei.ac.kr}           %  \\
           \and
           Hyunjung Shim \at
              \email{kateshim@yonsei.ac.kr}           %  \\
           \and
           $^1$\ School of Integrated Technology, Yonsei Institute of Convergence Technology, Yonsei University, Seoul, Republic of Korea.\\
           $^*$\ indicates a corresponding author and $^+$\ indicates an equal contribution.
}

\date{Received: 29 April 2019 / Accepted: 19 February 2020}

\maketitle

\begin{abstract}
Generative adversarial networks (GANs) successfully generate \comeng{high quality} data by learning a mapping from a latent vector to the data. Various studies assert that the latent space of a GAN is semantically meaningful and can be utilized for advanced data analysis and manipulation. To analyze the real data in the latent space of a GAN, it is necessary to build an inference mapping from the data to the latent vector. \comeng{This paper proposes} an effective algorithm to accurately infer the latent vector by utilizing \comeng{GAN discriminator features}. Our primary goal is to \comeng{increase inference mapping accuracy with} minimal training overhead. Furthermore, using the proposed algorithm, we suggest \comeng{a conditional} image generation algorithm, namely a spatially conditioned GAN. \comeng{Extensive} evaluations \comeng{confirmed} that the proposed inference algorithm \comeng{achieved} more semantically accurate inference mapping than existing methods and can be successfully applied to advanced conditional image generation tasks.
\end{abstract}
\section{Introduction} \label{intro}
Generative adversarial networks (GANs) have demonstrated remarkable progress in successfully reproducing real data distribution, particularly for natural images. Although GANs impose few constraints or assumptions on their model definition, they are capable of producing sharp and realistic images. To this end, training GANs \comeng{involves adversarial} competition between a generator and discriminator: the generator learns the generation process formulated by mapping from the latent distribution $P_\mathrm{z}$ to the data distribution $P_\mathrm{data}$; and the discriminator evaluates the generation quality by distinguishing generated images from real images. 
\cite{ref02} \comeng{formulated} the objective of this adversarial training using the minimax game
\begin{equation} \label{eq1}
    {\mathop{\mathrm{min}}_\mathrm{G}\ }{\mathop{\mathrm{max}}_\mathrm{D} }  \mathop{\mathbb{E}}_{x\sim \mathit{P}_{\mathrm{data}}}[\log(\mathrm{D}(x))] + \mathop{\mathbb{E}}_{z\sim \mathit{P}_{\mathrm{z}}}[\log(1-\mathrm{D}(\mathrm{G}(z))]\ ,
\end{equation}
\noindent where $\mathbb{E}$ denotes expectation; $\mathrm{G}$ and $\mathrm{D}$ are the generator and discriminator, respectively; and $z$ and $x$ are samples drawn from $P_{\mathrm{z}}$ and $P_{\mathrm{data}}$, respectively. Once the generator learns the mapping from the latent vector to the data (i.e., $z \rightarrow x$), it is possible to generate arbitrary data corresponding to randomly drawn $z$. Inspired by this pioneering work, various GAN models have been developed to improve training stability, image quality, and diversity of the generation.

In addition to image generation, GAN models are an attractive tool for building interpretable, disentangled representations. \comeng{Due} to their semantic power, several studies \citep{ref07,ref21} show that data augmentation or editing can be \comeng{achieved} by simple operations in the \comeng{GAN latent space}. To utilize the semantic representation derived by the \comeng{GAN latent space}, we \comeng{need to} establish inference mapping from the data to the latent vector (i.e., $x \rightarrow z$). \comeng{Previous studies generally adopt acyclic or cyclic inference mapping approaches to address the inference problem.}

\comeng{Acyclic} inference models develop inference mapping $x \rightarrow z$ independently from generation mapping (i.e., GAN training).
Consequently, learning this inference mapping can be formulated as minimizing image reconstruction error through latent optimization. \comeng{Previous} studies \citep{ref34,ref21} solve this optimization problem by finding \comeng{an} inverse generation mapping, $\mathrm{G}^{-1}\left (x \right)$, using a non-convex optimizer. However, \comeng{calculating this inverse path suffers from multiple local minima due to the generator's non-linear and highly complex nature;} thus it is \comeng{difficult} to reach the global optimum. In addition, \comeng{the consequentially} heavy computational load at runtime limits practical applications. To alleviate computational load at runtime, iGAN \citep{ref33} first proposed a hybrid approach, estimating from $x \rightarrow z^0$ and then $z^0 \rightarrow z$, where $z^0$ \comeng{is the initial state for} $z$. Specifically, iGAN \comeng{predicted} the initial latent vector for $x$ using an encoder model ($x \rightarrow z^0$), then \comeng{used} it as the initial optimizer value to compute the final estimate $z$ ($z^0 \rightarrow z$). Although the encoder model accelerates execution time \comeng{for} the testing phase, this initial estimate $x \rightarrow z^0$ is often inaccurate due to disadvantage of its encoder models, \comeng{and consequential} image reconstruction loss \comeng{presents} performance limitations that result in missing important attributes of the input data. \comeng{Section \ref{sec3_1}} presents a detailed discussion of various inference models.

\comeng{Cyclic} inference models \citep{ref05,ref06} consider bidirectional mapping, $x \leftrightarrow z$. That is to say, inference \comeng{learning} and generation mapping are considered simultaneously. \comeng{In contrast to} acyclic inference, cyclic inference aims to train the generator using feedback from inference mapping. For example, \citep{ref05,ref06} develop a cyclic inference mapping \comeng{to} alleviate the mode collapse problem. However, its performance is relatively poor in terms of both generation quality and inference accuracy, \comeng{which} leads to blurry images and the \comeng{consequential} poor inference results in inaccurate inference mapping.

\begin{figure*}[t!]
  \centering
    \includegraphics[width=0.9\linewidth]{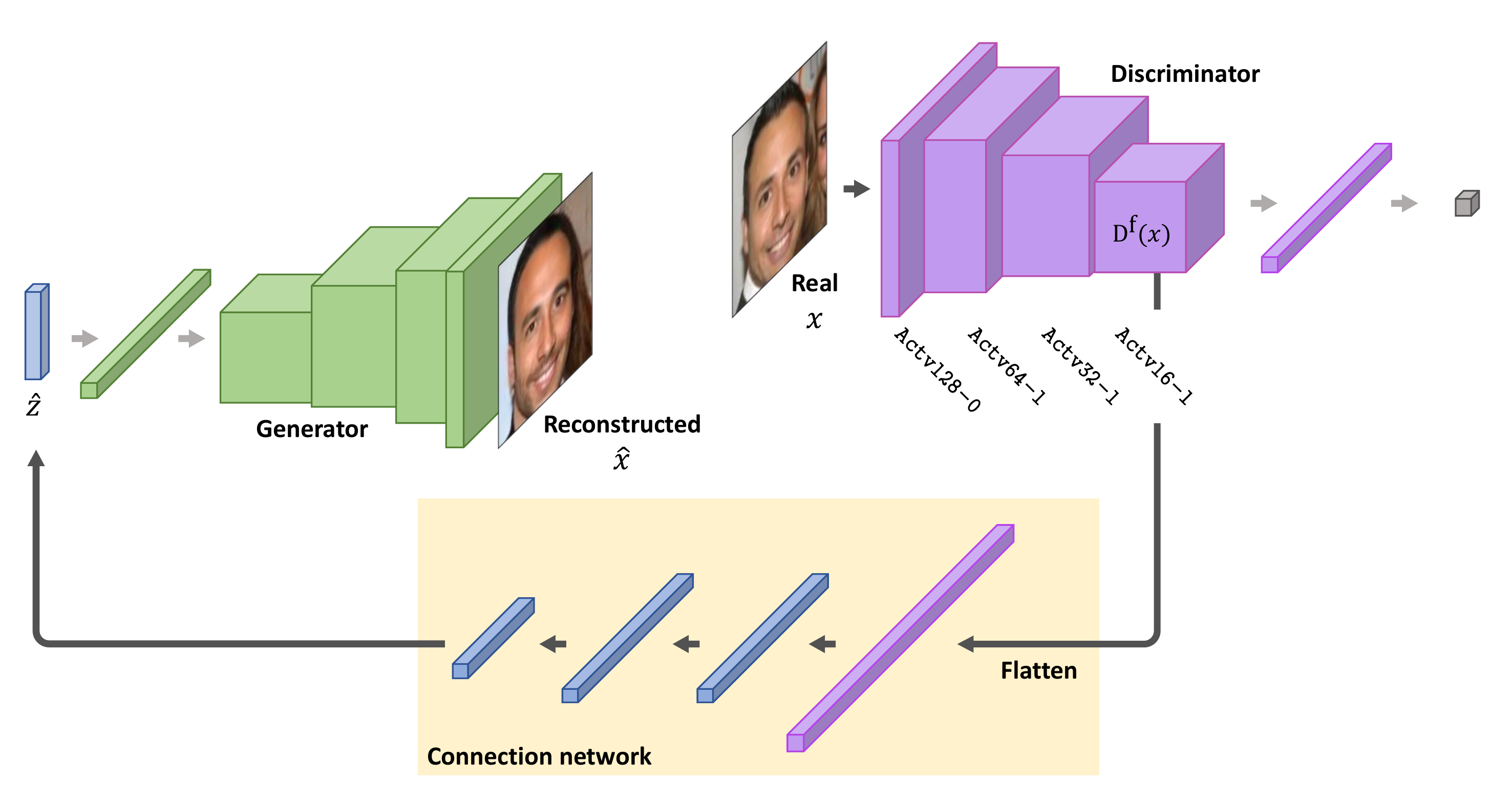}
    % \vskip -0.15in
    \caption{Network architecture for the proposed discriminator feature based inference (DFI) model, comprising a discriminator and connection network. The discriminator extracts feature ${\mathrm{D}^{\mathrm{f}}(x)}$ of input image $x$, and then the connection network infers the latent vector $\hat{z}$ of the input image. 
    }
    \label{figure_DFI}
\end{figure*}

\comeng{This paper proposes} a novel acyclic \comeng{discriminator feature based inference (DFI)} algorithm that exceeds both accuracy and efficiency of inference mapping for \comeng{current} techniques (Fig.~\ref{figure_DFI}). To improve inference accuracy, we suggest (1) replacing image reconstruction loss (evaluated with $x \sim {P}_{\mathrm{data}}$) with latent reconstruction loss (evaluated with $z \sim {P}_{\mathrm{z}}$) as an objective function for inference mapping, and (2) substituting the encoder with the discriminator as the feature extractor to prevent sample bias caused by latent reconstruction loss. \comeng{Section \ref{sec3_1}} discusses this issue in detail.

\comeng{Consequently, the proposed} algorithm performs inference in the order of $x \rightarrow \mathrm{D}^\mathrm{f}$ and then $ \mathrm{D}^\mathrm{f} \rightarrow z$, where $\mathrm{D}^\mathrm{f}$ implies the discriminator feature. Fortunately, since the pre-trained discriminator reveals $x \rightarrow \mathrm{D}^\mathrm{f}$, we only focus on finding $\mathrm{D}^\mathrm{f} \rightarrow z$. \comeng{Since} this mapping is a low-to-low dimensional translation, it is much more efficient than direct encoder based approaches of $x \rightarrow z$ in terms of model parameters. Thus, the proposed algorithm achieves computational efficiency in training.

\com{
\comeng{We need to consider two aspects to evaluate inference mapping}: how accurately the reconstructed image preserves semantic attributes, i.e., fidelity, and reconstructed image \comeng{quality} after applying the inference mapping. To quantify these two aspects, we \comeng{evaluated} inference models with five metrics: peak signal-to-noise ratio (PSNR), structural similarity index (SSIM), learned perceptual image patch similarity (LPIPS) \citep{zhang2018unreasonable}, face attribute classification accuracy, and Fr\'echet inception distance (FID)~\citep{dowson1982frechet}. We use multiple metrics for evaluation because no single metric is sufficient to quantify \comeng{both} aspects simultaneously. \comeng{The comparison confirmed} that the proposed DFI \comeng{outperformed} existing cyclic and acyclic inference in terms of both fidelity and quality.}

As a new and attractive application using the proposed inference mapping, \comeng{we developed} a spatially conditioned GAN (SCGAN) \comeng{that} can precisely control the spatial semantics for image generation. SCGAN successfully solves the spatially conditioned image generation \comeng{problem due to the accurate and efficient latent estimation from the proposed inference model.}

\comeng{Extensive} comparisons with \comeng{current} inference models and experimental analysis confirmed that \comeng{the proposed} inference algorithm \comeng{provided accurate} and efficient solutions for inference mapping.

\section{Preliminaries} \label{prelims}
The following sections describe \comeng{acyclic and cyclic inference models.}

\subsection{Acyclic inference models}
An acyclic inference model develops an inference mapping on top of a pre-trained GAN model. Thus, it consists of two steps.
\begin{enumerate} 
\item \comeng{Generation mapping is established by training a baseline GAN model.} 

\item \comeng{For inference mapping, the inference model is trained by minimizing the difference between $x$ and its reconstructed image $x'$, where $x'$ is $\mathrm{G}(z')$, $\mathrm{G}$ is determined at step (1), and $z'$ is the result of the inference model. }
\end{enumerate}
Since all generator and discriminator \comeng{parameters} are fixed during the inference mapping step, acyclic inference models leave baseline GAN performance intact.

CoGAN \citep{ref34} and BEGAN \citep{ref21} formulate inference mapping through a searching problem. Specifically, they search latent $z$, which is associated with the image most similar to target image $x$. \comeng{They use a} pixel-wise distance metric \comeng{to measure the similarity, and hence} this problem is defined as

\begin{equation} \label{eq2}
    {\mathop{\mathrm{min}}_{z} \ } \mathit{d} ( x  \  ,  \ \mathrm{G} (z) ), z \sim P_{\mathrm{z}}, z^0 \in \mathcal{R}^{|\mathrm{z}|},
\end{equation}
where $\mathit{d}(\cdot)$ is the distance metric and $z^0$ is \comeng{the} initial value \comeng{for} optimization. Eq. \ref{eq2} can be solved using advanced optimization algorithms, such as L-BFGS-B \citep{ref_BFGS} or Adam \citep{ref_adam}. Although this inference process is intuitive and simple, its results are often inaccurate and generally inefficient. \comeng{This} non-convex optimization easily falls into spurious local minima \comeng{due to the generator's non-linear and highly complex nature,} and estimation results are significantly biased by \comeng{the particular $z^0$ selected.} The optimization based inference algorithm \comeng{also} requires intensive computational effort in the testing phase, \comeng{which} is prohibitive \comeng{for} real-time applications.

To mitigate \comeng{these} drawbacks, iGAN \citep{ref33} focused on providing a good initial $z^0$ to assist \comeng{the optimization} search in terms of both effectiveness and efficiency, \comeng{proposing} a hybrid method \comeng{combining} an encoder model and optimization module sequentially. \comeng{The method} first predicts $z^0$ for the input $x$ using an encoder model, \comeng{and} the best estimate \comeng{for subsequent} $z$ is approximated by minimizing pixel difference between $\mathrm{G}(z)$ and $x$. \comeng{Thus,} the first step for training the encoder model $\mathrm{E}$ is defined as

\begin{equation} \label{eq3}
    {\mathop{\min}_\mathrm{\mathrm{E}} \ } \mathop{\mathbb{E}}_{x\sim P_{\mathrm{data}}} [ \mathit{d}( x  \  ,  \  \mathrm{G}  ( \mathrm{E} ( x ) ) ) ].
\end{equation}

The second step is the same \comeng{optimizing} Eq. \ref{eq2} except that the predicted latent vector is used as an initial value, $z^0 = \mathrm{E} (x)$. \comeng{Consequently}, iGAN reduces computational complexity for inference mapping at runtime. \com{However, since the encoder training utilizes samples from the data distribution, inference accuracy is severely degraded by the pre-trained generator having a mode missing problem, i.e., the generator is incapable of representing the minor modes. Section \ref{sec3_1} discusses this issue in more detail. Due to this accuracy issue, iGAN often misses important input data attributes, which are key components for interpreting the input.}

\subsection{Cyclic inference models}
Cyclic inference models learn inference and generation mapping \comeng{simultaneously}. 
\comeng{Variational (VAE) \citep{ref01} and adversarial (AAE) \citep{ref20} autoencoders} are popularly \comeng{employed} to learn bidirectional mapping between $z$ and $x$. Their model architectures are quite similar to autoencoders \citep{ref10}, \comeng{comprising} an encoder, i.e., the inverse generator, and a decoder, i.e., the generator. \comeng{In contrast to} autoencoders, VAE and AAE match latent distributions to prior distributions \citep{ref03}, enabling data generation. Whereas VAE utilizes Kullback-Leibler divergence to match latent \comeng{and} prior distributions, AAE utilizes adversarial learning for latent distribution matching. Although both algorithms establish bidirectional mapping between the latent and data distributions through stable training, their image quality is \comeng{poorer} than \comeng{for} unidirectional GANs. Specifically, generated images are blurry \comeng{with lost} details.

\comeng{The} ALI \citep{ref05} and BiGAN \citep{ref06} bidirectional GANs jointly learn bidirectional mapping between $z$ and $x$ in an unsupervised manner. They use \comeng{a} generator \comeng{to} construct forward mapping from $z$ to $x$, and then \comeng{an} encoder to model inference mapping from $x$ to $z$. To train the generator and the encoder simultaneously, they define a new objective function for the discriminator \comeng{to} distinguish the joint distribution, $\{\mathrm{G}(\mathit{z}), z\}$, from $\{x, \mathrm{E}(x)\}$. \comeng{Thus,} the ALI and BiGAN \comeng{objective function} is 
\begin{equation}
    {\mathop{\min}_{\mathrm{G}} }{\mathop{\max}_{\mathrm{D}} } \mathop{\mathbb{E}}_{x\sim P_{\mathrm{data}}} [\log( \mathrm{D}(x, \mathrm{E}(x))]     + \mathop{\mathbb{E}}_{z\sim P_{\mathrm{z}}}[\log(1-\mathrm{D}(\mathrm{G}(z), z)] .
\end{equation}

Although these models can reconstruct the original image from the estimated latent vector, generation quality is \comeng{poorer} than that \comeng{for} unidirectional GANs \comeng{due to} convergence issues \citep{ref37}. \comeng{In contrast,} they alleviate the \comeng{unidirectional GAN} mode collapse problem by utilizing inference mapping.

\comeng{The} VEEGAN \citep{ref09} and ALICE \citep{ref37} introduce an additional constraint that enforces the reconstructed image (or the latent vector) computed from the estimated latent vector (or image) to match the original image (or latent vector). This improves either mode collapse or training instability for bidirectional GANs. Specifically, VEEGAN utilizes cross-entropy between $P_\mathrm{z}$ and $\mathrm{E}(x)$, defined as the reconstruction penalty in the latent space, \comeng{to establish} joint distribution matching; \comeng{whereas} ALICE aims to improve GAN training instability by adopting conditional entropy, defined as cycle consistency \citep{ref38}. Although both methods improve joint distribution matching \comeng{performance}, they still suffer from \comeng{discrepancies} between theoretical optimum and practical convergence \citep{ref37}, resulting in either slightly blurred generated images or inaccurate inference mapping.

\section{Discriminator feature based inference} \label{proposed_dfi}

\comeng{The} proposed algorithm \comeng{is an} acyclic inference model, \comeng{in that} the training process is isolated from GAN training, i.e., both the generator and discriminator are updated. This implies that baseline GAN model \comeng{performance} is not affected by inference mapping. Our goal \comeng{with the proposed pre-trained GAN model}, is to (1) increase inference mapping accuracy and (2) build a real-time inference algorithm with minimal training overhead.

\comeng{Therefore,} we propose a discriminator feature based inference algorithm to achieve these goals. Specifically, we build a connection network that establishes the mapping from image features to the latent vector by minimizing latent reconstruction loss. We formulate the objective for learning the connection network as 

\begin{equation}\label{eq4}
    {\mathop{\mathrm{min}}_\mathrm{CN} \ }
    \mathop{\mathbb{E}}_{z\sim P_{\mathrm{z}}} [\mathit{d} ( z,\  \mathrm{CN} ( \mathrm{D^f} ( \mathrm{G}  (z) ) ))] ,
\end{equation}
 where $\mathrm{CN}$ is the connection network, \comeng{and} $\mathrm{D^f}(x)$ indicates the discriminator feature vector of $x$, extracted from the last layer of the discriminator.
 
 In our framework, the generated image from $z$ is projected onto the discriminator feature space, and this feature vector then maps to the original $z$ using the connection network. It is important to understand that correspondences between the latent vector $z$ and discriminator features $\mathrm{D^f}(x)$ are automatically set for arbitrary $z$ once both generator and discriminator \comeng{training} ends. Hence, the connection network is trained to minimize the difference between $z$ and its reconstruction by the connection network.

\comeng{The} following sections provide the rationale for the proposed algorithm (Section \ref{sec3_1}), suggest a new metric for inference mapping (Section \ref{selfFID}), and then introduce a spatially conditioned GAN (SCGAN) practical application of \comeng{the proposed} DFI (Section \ref{sec:SCGAN}). We stress that SCGAN addresses spatial conditioning for image generation for the first time.

\subsection{Rationale } \label{sec3_1}
\noindent \textbf{Why DFI \comeng{is superior to previous} acyclic algorithms.}
\comeng{The classic iGAN acyclic inference algorithm} uses an encoder based inference model that minimizes image reconstruction loss in Eq. \ref{eq3} in the first stage. \comeng{In contrast, the proposed} DFI aims to minimize latent reconstruction loss for training the connection network. \comeng{These approaches are identica}l for an ideal GAN, i.e., perfect mapping from $z$ to $x$. However, \comeng{practical} GANs notoriously suffer from mode collapse; \comeng{where} the generator only covers a few major modes, \comeng{ignoring} the often many minor modes.

Suppose that the distribution reproduced by the generator $P_{\mathrm{g}}$ does not cover the entire distribution of $P_{\mathrm{data}}$, i.e., mode collapse. Then, consider the sample \com{$x$, where $P_\mathrm{g}(x)=0$ and $P_{\mathrm{data}}(x) \neq 0$}. For such a sample, image reconstruction loss between $x$ and \com{$x' = \mathrm{G}(\mathrm{E}(x))$} by Eq.~\ref{eq3} is ill-specified~\citep{ref09}, \com{where $\mathrm{E}$ is an inference algorithm that maps an image to a latent vector,} \comeng{since} $x'$ is undefined by the generator. \comeng{Any} inference model trained with image reconstruction loss \comeng{inevitably} leads to inaccurate inference mapping, \comeng{due to} those undefined samples. \com{In other words, the image reconstruction suffers from noisy annotations since it learns the mapping from the real image to its latent code, which are latent codes for real images not covered by the generator. This leads to inference accuracy degradation, e.g. attribute losses and blurry images.}

\comeng{In contrast}, latent reconstruction loss only considers the mapping from \com{$z' = \mathrm{E}(\mathrm{G}(z))$ to $z \sim P_\mathrm{z}$}, i.e.,  latent reconstruction loss does not handle samples not covered by the generator. Thus, Eq.~\ref{eq4} solves a well-specified problem: \com{a set of accurate image-annotation pairs are used for training.} \comeng{This} can significantly influence inference accuracy, \comeng{and is} critical for acyclic inference models developed with a pre-trained generator having practical limitations, such as mode collapse.

\com{We stress that inference mapping using a fixed generator is trained via a set of image-latent pairs in a fully supervised manner. Since supervised learning performance largely depends on annotation quality, refining the training dataset to improve annotation accuracy often improves overall performance. In this regard, the proposed latent reconstruction loss can be interpreted as the improving annotation quality, because it tends to train inference mapping using correct image-latent pairs.}

\vskip 0.15in

\noindent \textbf{Why the discriminator is a good feature extractor for DFI.}
Although the discriminator is typically abandoned after GAN training, we claim it is a useful feature extractor for learning the connection network. The previous study \citep{ref07} empirically showed that discriminator features are powerful representations for solving general classification tasks. The discriminator feature representation is even more powerful \comeng{for inference mapping}, for the following reasons. 

To train the connection network using latent reconstruction loss, all training samples are fake samples, drawn from $z \sim P_\mathrm{z}$, as described in Eq. \ref{eq4}. Although utilizing latent reconstruction loss is useful to construct a well-specified problem, this naturally leads to sample bias, i.e., a lack of real samples, $x \sim {P}_{\mathrm{data}}$, during training. To mitigate training bias, we utilize the discriminator as a feature extractor, because the discriminator feature space already provides comprehensive representation for both real and fake samples. \comeng{Thus,} the pre-trained discriminator learns to classify real and fake samples during training. Consequently, we expect that the \comeng{discriminator} feature space can bridge the discrepancy between real and fake samples, \comeng{helping to} alleviate sample bias.

\subsection{Metrics for assessing inference accuracy}\label{selfFID}

\comeng{Although} several metrics are available for evaluating GAN models, an objective metric for assessing inference models has not been \comeng{established}. Developing a fair metric is beneficial to encourage constructive competition, \comeng{and hence} escalate the advance of inference algorithms. 

Two aspects should be considered to evaluate inference algorithm accuracy: semantic fidelity and reconstructed image quality. We utilize LPIPS~\citep{zhang2018unreasonable} and face attribute classification (FAC) accuracy~\citep{liu2019stgan} to measure reconstructed image semantic fidelity, i.e., similarity to the original image. Section~\ref{Compare_acyclic} empirically discusses the high correlation between LPIPS and FAC accuracy. Therefore, we employ LPIPS as the measure for semantic fidelity for further experiments because FAC accuracy is not flexible enough to apply on various datasets. \comrevision{In addition, We suggest FID~\citep{dowson1982frechet} to measure the image quality, i.e. how realistic the image is. We emphasize that LPIPS is more suitable to measure the fidelity of the reconstructed image while FID is more suitable to measure the image quality of the reconstructed image. }

\noindent \com{\textbf{LPIPS} The learned perceptual image patch similarity (LPIPS) metric for image similarity utilizes a pre-trained image classification network e.g. AlexNet~\citep{krizhevsky2014one}, VGG~\citep{ref35}, and SqueezeNet~\citep{iandola2016squeezenet}) to measure feature activation differences between two images, and returns a similarity score using learned linear weights. LPIPS can capture semantic fidelity because both low and high level features of the pre-trained network influence similarity.}

\noindent \textbf{FID} \com{Although LPIPS is a powerful metric for semantic fidelity, it does not reflect reconstructed image quality. We need to consider whether the reconstructed image is on the image manifold to measure quality}. FID is a popular metric that quantifies sample quality and diversity for generative models, particularly GANs \citep{lucic2018gans, zhang2018self, brock2018large}, where smaller FID score indicates fake samples have (1) high quality (i.e., they are sharp and realistic) and (2) various modes similar to real data distribution.

FID represents the Fr\'echet distance~\citep{dowson1982frechet} between the moments of two Gaussians, representing the feature distribution of real images and randomly drawn fake images. \comrevision{We also utilize FID for evaluating inference algorithms. For that, the Fr\'echet distance between moments of two Gaussians are measured where two Gaussians represent feature distributions for real images and their reconstructed images.}

\comrevision{The FID for the inference algorithm} can be expressed as
\begin{equation}
\begin{split}
    d^2 (\mathrm{(\mu, \Sigma), (\mu_{R}, \Sigma_{R})) = ||\mu - \mu_{R}||_2^2} \  + \\ \mathrm{Tr}(\mathrm{\Sigma+\Sigma_{R} - 2(\Sigma\Sigma_{R})^{1/2}}),
\end{split}
\end{equation}
where $\mathrm{(\mu, \Sigma)}$ (or $\mathrm{(\mu_{R}, \Sigma_{R})}$) indicates the mean vector and covariance matrix for the Inception features computed from real images (or reconstructed images obtained by inference mapping).

\begin{figure}[t!]
  \centering
    \includegraphics[width=0.95\columnwidth]{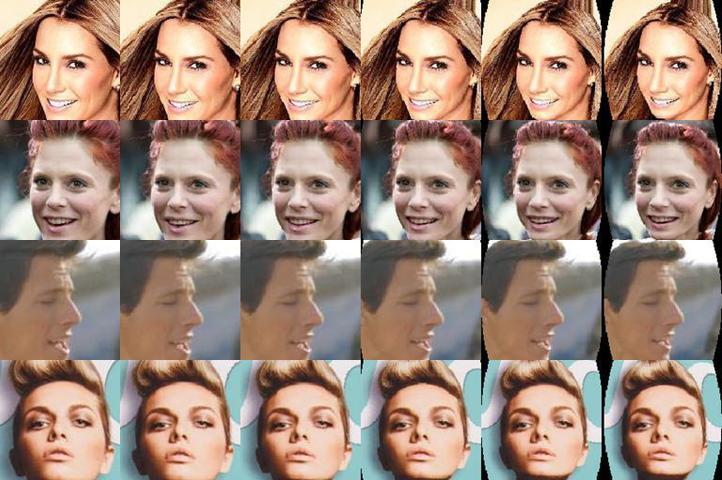}
    % \vskip -0.15in
    \caption{\comrevision{Fish-eye distortion examples. Each column depicts distorted results with distortion coefficient 0.0 (original), 0.1, 0.2, 0.3, 0.4, and 0.5.}}
    \label{figure_metric_fisheye_example}
    %\vskip -0.2in
\end{figure}

\begin{figure}[t!]
  \centering
    \includegraphics[width=0.95\columnwidth]{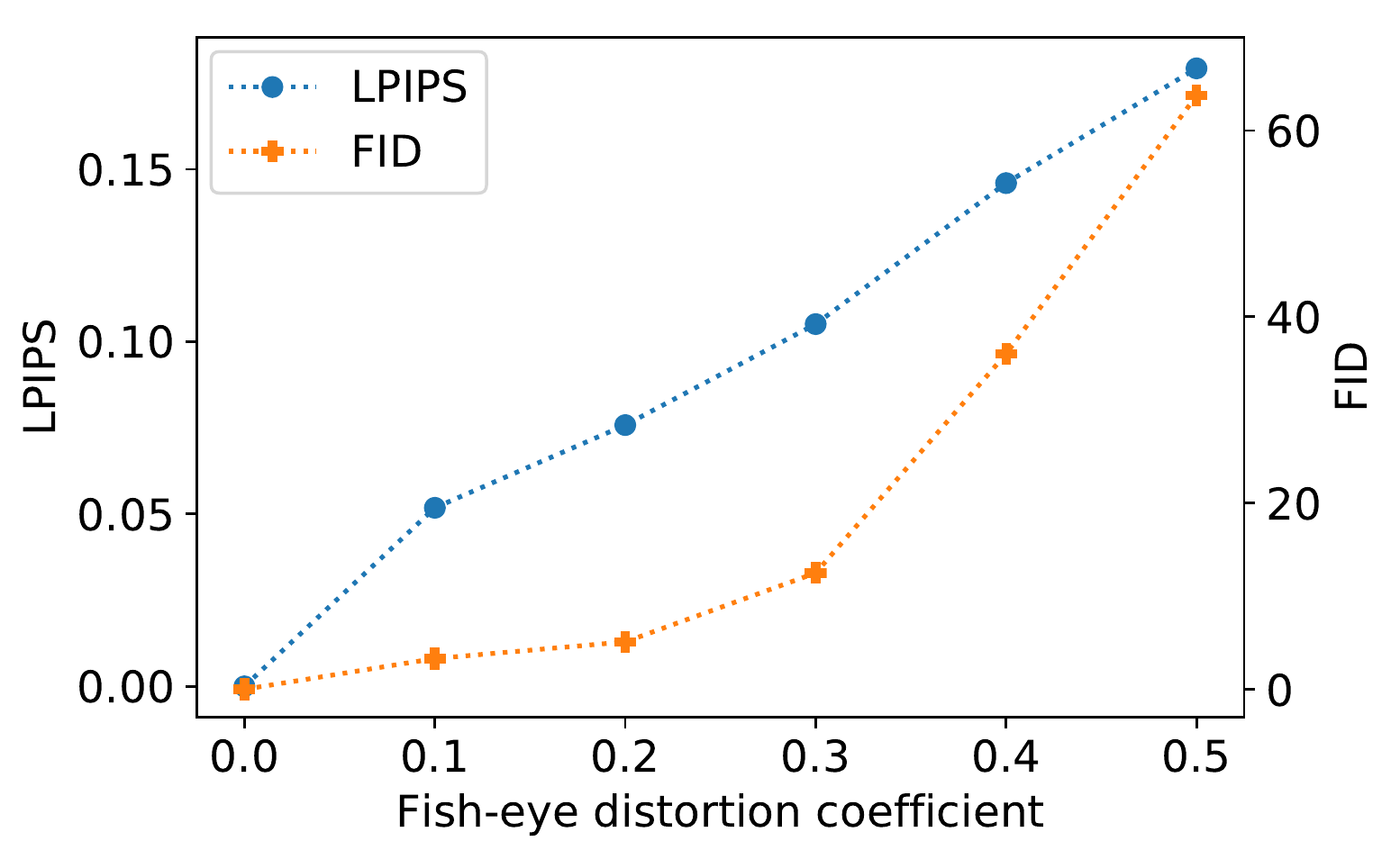}
    % \vskip -0.15in
    \caption{\comrevision{Quantitative comparison between LPIPS and FID.  Each score is computed using the real images and their distorted images where the fish-eye distortion coefficient gradually increases from 0 to 0.5.}}
    \label{figure_metric_fisheye}
    %\vskip -0.2in
\end{figure}

\begin{figure}[t!]
  \centering
    \includegraphics[width=0.95\columnwidth]{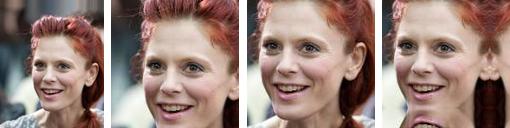}
    % \vskip -0.15in
    \caption{\comrevision{Translation examples with two different padding strategies. First image is an original image. From the second to fourth, each image depicts the result of translation (0, 0), (21, 21) with raw padding, and (21, 21) with reflection padding respectively.}}
    \label{figure_metric_translation_example}
    %\vskip -0.2in
\end{figure}

\begin{figure}[t!]
  \centering
    \includegraphics[width=0.95\columnwidth]{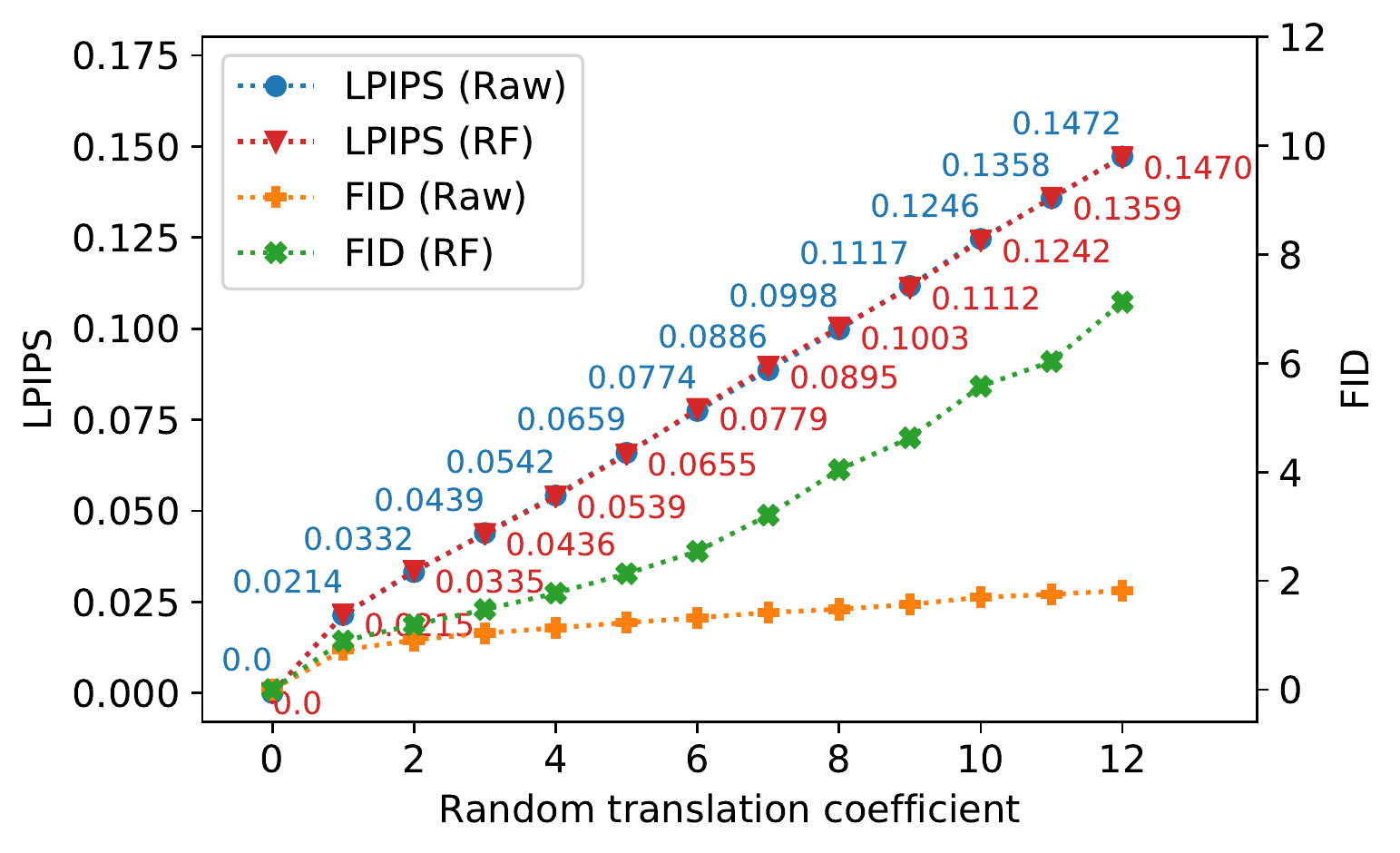}
    % \vskip -0.15in
    \caption{\comrevision{Quantitative results of LPIPS and FID score between the real images and their translated images according to the random translation coefficient. To clearly visualize the difference in LPIPS scores for two padding strategies, LPIPS scores are overlaid to each point.}}
    \label{figure_metric_translation}
    %\vskip -0.2in
\end{figure}

\comrevision{It is important to note that the FID for the inference algorithm is an unbiased estimator since each reconstructed image has its real image pair. Thus, the FID for the inference algorithm provides a deterministic score for given real image set, reliable even for small test samples. }

\vskip 0.1in
\noindent \comrevision{\textbf{Rationale of using both metrics.} To justify the above mentioned properties of LPIPS and FID, we provide one exemplar case and two empirical studies. First, the advantage of LPIPS can be clearly demonstrated by the following example. Note that LPIPS guarantees the ideal reconstruction if its score is zero. Meanwhile, any permutation of perfect reconstruction can yield zero FID. This indicates that LPIPS is reliable to measure faithful reconstruction; FID is not.}

\comrevision{Contrary, LPIPS is overly sensitive to structural perturbations between the two images, thus not suitable to assess the general image quality. In fact, such a sensitivity is natural because LPIPS directly measures the pixel-level difference between two feature activations across all scales. It should be noted that FID is robust against the structural perturbations because it does not evaluate the pixel-level difference between the feature maps of the two images, but evaluates the statistical differences of the two high-level feature distributions. To demonstrate the advantage of FID, we carry out two experiments; measuring LPIPS and FID between (1) the real images and their fish-eye distorted images, and (2) the real images and their translated images. The experiment utilizing fish-eye distortions is also conducted in \cite{zhang2018unreasonable}. Figure~\ref{figure_metric_fisheye_example} depicts several distorted images. From the left to the right, the fish-eye distortion parameter increases(the larger the parameters, the harsher the distortion). Figure~\ref{figure_metric_fisheye} shows LPIPS and FID scores when distortion parameters increases. We observe that FID does not change much for the images with small distortions while the score exponentially increase for the images with large distortions. This makes sense and is analogous to how human evaluates the difference between the two images; the three images corresponding to small distortions in Figure~\ref{figure_metric_fisheye} (parameter 0.1, 0.2 and 0.3) are more similar to the original while the last two images (parameter 0.4 and 0.5) are clearly different from the original. Unlike FID, LPIPS are linearly increases as the distortion parameter increases. That means, LPIPS is not robust against small structural perturbations.}

\comrevision{We further investigate the property of FID and LPIPS by applying random translation in real images. For padding after translation, we select two strategies; raw padding and reflection padding. For raw padding, we center crop image after shifting the original real image. For reflection padding, we center-crop image first and shift the cropped image with reflection padding. As seen from Figure~\ref{figure_metric_translation_example}, raw padding results in realistic images whereas reflection padding creates creepy and unrealistic faces. We apply random shift for both vertical and horizontal axis of the image within the range $(-t, -t) \sim (t, t)$ where $t$ is a translation coefficient. Figure~\ref{figure_metric_translation} describes LPIPS and FID score as the translation coefficient increases. Interestingly, we observe that the difference between LPIPS scores for the two padding strategies are marginal. Contrary, the difference between the two FID scores for the two different padding strategies is considerable. Specifically, the translation using raw padding leads extremely small FID scores (FID less than 2 is almost negligible) while the translation using reflection padding yields meaningful difference in FID scores. These results present that the FID is more suitable to measure image quality, i.e., how realistic the generated samples are, than LPIPS.}

\comrevision{From two empirical studies, we conclude that FID is more robust to small structural perturbations in images than LPIPS. Owing to this attractive properties, we confirm that FID better evaluates the image quality than LPIPS. Considering the advantages of FID and LPIPS in different aspects, we claim that both FID and LPIPS should be used for assessing inference algorithms. For this reason, we report both scores as quantitative measures for various inference algorithms.}

\vskip 0.15in

\com{Although we include PSNR and SSIM metrics, their scores do not reflect perceptual quality well. We argue that LPIPS and FID can better assess inference algorithm modeling power. Section~\ref{Compare_acyclic} empirically shows PSNR and SSIM demerits as accuracy measures for inference algorithms.}

\subsection{Spatially conditioned image generation} \label{sec:SCGAN}

\begin{figure*}[t!]
  \centering
    \includegraphics[width=0.8\linewidth]{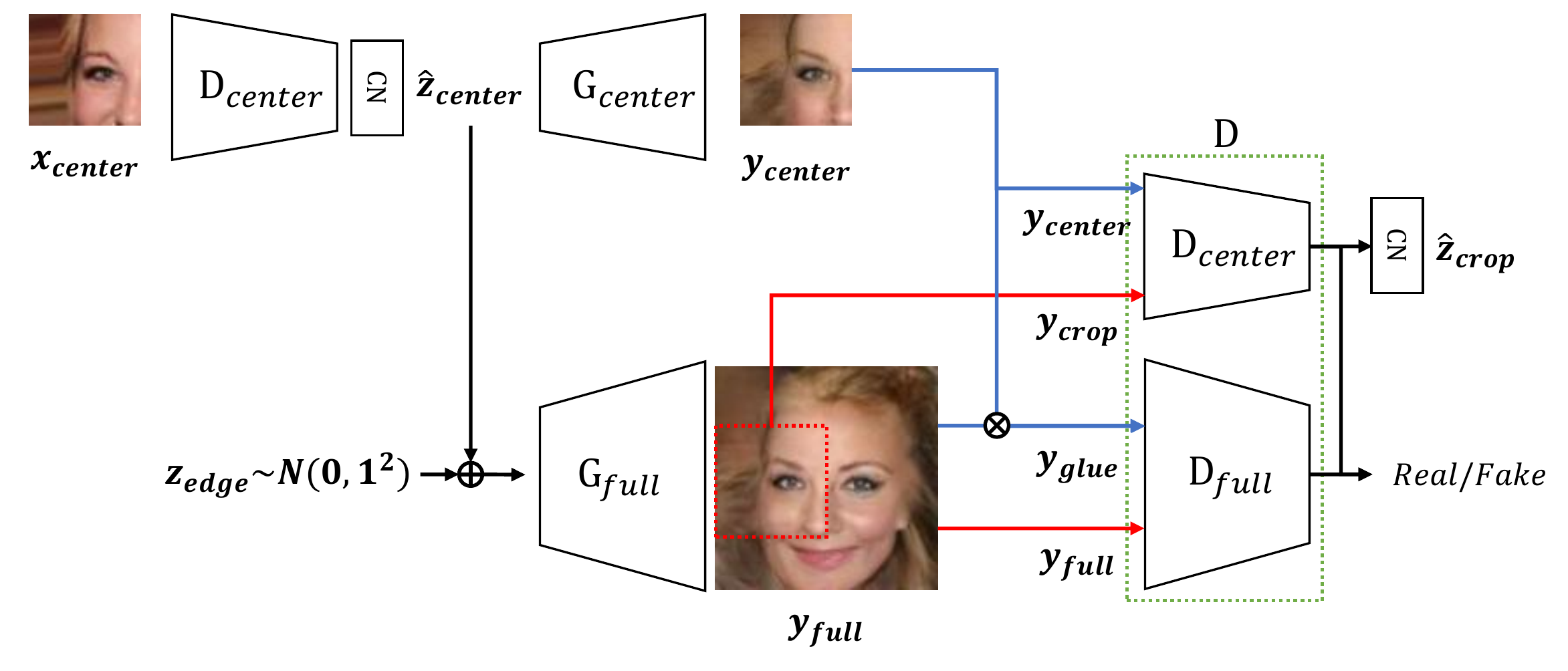}
    % \vskip -0.15in
    \caption{Network architecture for the proposed SCGAN. $\bigoplus$ denotes concatenation of latent vectors. $\bigotimes$ denotes image replacement. ${y}_{glue}$ is identical to ${y}_{full}$ except the image center (square area outlined by red dots), \comeng{which} was replaced with ${y}_{center}$. The design choice for $\mathrm{D}$ \comeng{was} motivated by ~\cite{ref40}, \comeng{and} includes a global discriminator$\mathrm{D}_{full}$ and local discriminator $\mathrm{D}_{center}$.}
    \label{figure_SCGAN}
    %\vskip -0.2in
\end{figure*}

Semantic features are key components for understanding and reflecting human intentions because they are closely related to human interpretation. Indeed, the way humans define tasks is never specific but is \comeng{rather} abstract or only describes semantic characteristics. For example, \comeng{human facial} memorizing does not rely on local details, such as skin color or roughness, but focuses more on facial shape, hair color, presence of eyeglasses\comeng{, etc}. Therefore, from the human \comeng{viewpoint}, useful image analysis and manipulation should be associated with extracting semantic attributes of the data and modifying them effectively. \comeng{Since} the proposed inference algorithm developed by the connection network establishes semantically accurate inference mapping, \comeng{combining this} inference algorithm with standard GANs can \comeng{provide} strong baseline models for data manipulation and analysis applications.

\comeng{Therefore}, we suggest a new conditional image generation algorithm: spatially conditioned GAN (SCGAN). SCGAN extracts the latent vector of input using \comeng{the proposed} inference algorithm and uses it for spatially conditioned image generation.

\comeng{In particular}, we specify the position input image \comeng{position}, and then generate the surroundings using SCGAN. In this process, the generated surrounding region should naturally and seamlessly match the input image. Among \comeng{the infinite methods to generate the outside regions}, our goal is to achieve semantically seamless results. \comeng{Therefore}, SCGAN first maps the input image to its latent vector \comeng{using DFI}, which encodes the semantic attributes. Given the latent vector of input, spatially conditioned image generation is conducted by generating the large image (full size) \comeng{such that} the image region at the input position is the reconstructed input and its surroundings are newly generated. The generated surroundings should seamlessly match the semantics of the input with \comeng{reasonably} visual quality. \comeng{Since} many possible surroundings can match the input, we formulate the latent vector of the generated image by concatenating the random vector with the latent vector of input. \comeng{Thus}, SCGAN maintains \comeng{input} semantic attributes \comeng{while} allowing diverse image surroundings.

Figure~\ref{figure_SCGAN} illustrates the \comeng{proposed SCGAN} architecture. To extract the latent vector for input image ${x}_{center}$, we first train baseline GANs, \comeng{comprising} a generator ${\mathrm{G}}_{center}$ and discriminator ${\mathrm{D}}_{center}$, \comeng{and then} fix the GANs and train the connection network ($\mathrm{CN}$) to utilize DFI. Given the fixed ${\mathrm{D}}_{center}$ and $\mathrm{CN}$, we compute ${\hat{z}}_{center}$, the estimated latent vector for ${x}_{center}$. To account for diverse surroundings, \comeng{we concatenate} a random latent vector ${z}_{edge}$ with ${\hat{z}}_{center}$ and feed this into the generator ${\mathrm{G}}_{full}$. This network learns to map the concatenated latent vector to full size image ${y}_{full}$, which is the final output image. 

We \comeng{train} ${\mathrm{G}}_{full}$ to satisfy ${y}_{crop}$: the image center of ${y}_{full}$ should reconstruct ${x}_{center}$; and ${y}_{full}$ should have a diverse boundary region and sufficiently high overall quality. To meet the first objective, the na{\"i}ve solution is to minimize L1/L2 distance between ${y}_{crop}$ and ${x}_{center}$. However, as reported previously~\citep{larsen2015autoencoding}, combining image-level loss with adversarial loss increases GAN training \comeng{instability}, resulting in quality degradation. Hence, we define reconstruction loss in the latent space, i.e., we map ${y}_{crop}$ onto its latent vector via DFI (${\mathrm{D}}_{center}$ and $\mathrm{CN}$), then force it to match ${\hat{z}}_{center}$. \comeng{Thus}, the semantic similarity between the input and its reconstruction is preserved.

\comeng{To ensure} seamless composition between reconstructed and generated regions, adversarial loss for ${\mathrm{G}}_{full}$ consists of feedback from ${y}_{full}$ and ${y}_{glue}$. ${y}_{glue}$ is obtained by substituting the generated image center ${y}_{crop}$ with the reconstructed input ${y}_{center}$. This term for ${y}_{glue}$ helps generate visually pleasing images, i.e., reconstructed input and its surroundings are seamlessly matched.
\comeng{Thus,} generator loss includes two adversarial losses and latent reconstruction loss, 

\begin{equation}
    {\mathop{\mathrm{min}}_{\mathrm{G}_{full}} \ }  0.5 \ {L}_\mathrm{G}^{adv} \ + \alpha \ {L}^{recon},
\end{equation}

\[{L}^{recon} = \left \| \hat{z}_{center} \ - \ \mathrm{CN}(\mathrm{D}_{center}({y}_{crop})) \right \|_{1}, \]
\comeng{and}
\[{L}_{\mathrm{G}}^{adv} = \mathop{\mathbb{E}}_{z_{edge}\sim \mathit{P}_{\mathrm{z}}} [\mathrm{log}(1-\mathrm{D}(y_{full})) + \\ \mathrm{log}(1- \mathrm{D}({y}_{glue}))],\]
\comeng{respectively.}

Semantic consistency between reconstructed and generated regions is important to create natural images. To obtain locally and globally consistent images, we utilize local and global discriminator $\mathrm{D}$ ~\citep{ref40} architecture that uses discriminator features from both ${\mathrm{D}}_{center}$ and ${\mathrm{D}}_{full}$. \comeng{We also} employ PatchGAN \citep{ref42} architecture to strengthen the discriminator, accounting for semantic information from patches in the input, and apply the zero-centered gradient penalty (0GP) \citep{ref41} to ${D}_{full}$ to facilitate high resolution image generation. Considering adversarial loss and zero-centered gradient penalty, discriminator loss \comeng{can be expressed as}

\begin{equation}
{\mathop{\mathrm{max}}_{\mathrm{D}_{full}} \ }  {L}_\mathrm{D}^{adv} + 0.5\ {L}_\mathrm{G}^{adv} \ + \ {L}^{GP}
\end{equation}
\[{L}_\mathrm{D}^{adv} = \mathop{\mathbb{E}}_{x\sim \mathit{P}_{\mathrm{data}}} [\mathrm{log}\mathrm{D}(x)], \ 
and
{L}^{GP} = \frac{\gamma }{2} \mathop{\mathbb{E}}_{x\sim \mathit{P}_{\mathrm{data}}} [\left \| \nabla \mathrm{D}(x) \right \| ^2].\]
\section{Experimental results}\label{evaluation}

\begin{table*}[t!] 

  \centering
  \resizebox{\textwidth}{!}{
    \begin{tabular}{c | c c | c c | c}
    \toprule
        {} & \multicolumn{2}{c|}{Training loss} & \multicolumn{2}{c|}{Architecture} & {} \\
        {} & {\makecell{Image reconstruction loss}} & {\makecell{Latent reconstruction loss}} & \makecell{Encoder trained \\ from scratch} & \makecell{Discriminator ($\mathrm{D^f}$) \\ + $\mathrm{CN}$ network} & {\makecell{Additional\\optimization}} \\
        \midrule
        {$\mathrm{ENC_{image}}$} & {\checkmark} & {} & {\checkmark} & {} & {} \\
        {$\mathrm{ENC^{opt}_{image}}$ (iGAN)} & {\checkmark} & {} & {\checkmark} & {} & {\checkmark} \\
        {$\mathrm{ENC^{}_{latent}}$} & {} & {\checkmark} & {\checkmark} & {} & {} \\
        {$\mathrm{ENC^{opt}_{latent}}$} & {} & {\checkmark} & {\checkmark} & {} & {\checkmark} \\
        {$\mathrm{DFI_{image}}$} & {\checkmark} & {} & {} & {\checkmark} & {} \\
        {$\mathrm{DFI^{opt}_{image}}$} & {\checkmark} & {} & {} & {\checkmark} & {\checkmark} \\
        {$\mathrm{DFI}$ (proposed)} & {} & {\checkmark} & {} & {\checkmark} & {} \\
        {$\mathrm{DFI^{opt}}$} & {} & {\checkmark} & {} & {\checkmark} & {\checkmark} \\
        
        \bottomrule
    \end{tabular}
    }
    \vskip 0.1in
    \caption{\comrevision{The abbreviation for various baseline models and the variants of the proposed models.}}
  \label{table_terms}
  \vskip -0.2in
\end{table*}

\comrevision{For a concise expression, we use the abbreviation for network combinations for the rest of the paper. Table~\ref{table_terms} summarizes the component of each network model and its abbreviation. For additional optimization, each baseline model first infers initial $z_0$ and then optimize $z$ by following Eq.~\ref{eq2} for 50 iterations~\citep{ref33}.}

\vskip 0.1in
\noindent \textbf{Metrics for quantitative evaluation.}
\com{\comeng{We employed} PSNR, SSIM, LPIPS, face attribute classification (FAC) accuracy, FID, and a user study} \comeng{to quantitatively evaluate various inference algorithms.} For the user study, 150 participants compared real images with their reconstruction from all inference models to select the one that most similar to the real image. Each participant \comeng{then responded} to three questions. 
\begin{enumerate}
\item \comeng{We provided} seven images: the original and reconstructed images from \comrevision{(a) $\mathrm{ENC_{image}}$, (b) $\mathrm{ENC^{opt}_{image}}$ (iGAN), (c) $\mathrm{ENC^{}_{latent}}$, (d) $\mathrm{ENC^{opt}_{latent}}$, (e) $\mathrm{DFI}$, and (f) $\mathrm{DFI^{opt}}$.} We asked the participant to select the image \comeng{most similar to the original image} from among the six reconstructed images. 
\item \comeng{The participant was asked to explain the} reason for their choice. 
\item We provided \comrevision{$\mathrm{DFI}$ and $\mathrm{DFI-VGG16}$} (discussed in Section~\ref{subsction_dfi_vgg}) \comeng{images, and asked} participants to select the one most similar to the original. 
\end{enumerate}
This \comeng{was} repeated 25 times using different input images.

\vskip 0.1in
\noindent \textbf{State-of-the-art inference algorithms for comparison.}
Experimental comparisons are conducted for acyclic and cyclic inference models. First, we compare the proposed inference algorithm with three acyclic inference algorithms: na{\"i}ve encoder \comrevision{($\mathrm{ENC_{image}}$ and $\mathrm{ENC_{latent}}$), hybrid inference by iGAN~\citep{ref33} ($\mathrm{ENC^{opt}_{image}}$), and hybrid inference combined with DFI ($\mathrm{DFI^{opt}}$).} \comeng{The proposed DFI model outperformed all three acyclic models for all} four evaluation methods (LPIPS, FAC accuracy, FID, and user study).

\comeng{We then compared current} cyclic models (VAE, ALI/BiGAN, and ALICE) with \comeng{the proposed DFI based model} upon various baseline GAN models. \comeng{Cyclic model inference mapping} influences baseline GAN performance, whereas \comeng{acyclic model (i.e., DFI)} inference mapping does not.
\comeng{We combined} six different baseline GANs with DFI \comeng{for this evaluation}: DCGAN \citep{ref07}, LSGAN \citep{ref11}, DFM \citep{ref13}, RFGAN \citep{ref19}, SNGAN \citep{ref43}, and WGAN-GP \citep{ref12}. \comeng{These six} were selected because they are significantly different from each other in terms of loss functions or network architectures. We evaluated all results with $\le (64, 64, 3)$ resolution since cyclic models are unstable for high resolution images. \comeng{To illustrate} DFI scalability, we build inference mapping with high resolution GANs \citep{ref41, ref43} combined with DFI, and observed similar tendency in terms of inference accuracy for $(128, 128, 3)$ resolution images.

\vskip 0.1in
 \noindent \com{\textbf{Qualitative evaluation for DFI.} 
Generators learn rich linear structure in representation space due to the power of semantic representations of GAN latent space \citep{ref07}. To qualitatively evaluate semantic accuracy for the proposed DFI, we conducted two simple image manipulation tasks: latent space walking and vector arithmetic.}

\vskip 0.1in
\noindent \textbf{Model architecture for fair comparison.}
\comeng{To ensure} fair evaluation, \comeng{we based} baseline GAN \comeng{architectures on DCGAN} for low resolution and SNGAN for high resolution experiments, i.e., number of layers, filter size, hyper-parameters, etc. The connection network included just two fully connected (FC) layers: 1024 -- group normalization (GN) \citep{ref_GN} -- leaky rectified linear unit (Leaky ReLU) -- 1024 FC -- GN -- Leaky ReLU -- dimension of $P_\mathrm{z}$ FC.

\noindent \textbf{Datasets.}
One synthetic and three real datasets \comeng{were} used \comeng{for both qualitative and quantitative evaluations}. \comeng{We generated} eight Gaussian spreads \comeng{for the synthetic dataset distribution}. \comeng{Real datasets included} Fashion MNIST \citep{ref25}, CIFAR10 \citep{ref26}, and CelebA \citep{ref14}, \comeng{and were} all normalized on $[-1, 1]$. \comeng{Input dimensionality for} Fashion MNIST $=(28, 28, 1)$; CIFAR10 $=(32, 32, 3)$; and CelebA $=(64, 64, 3)$ and $(128, 128, 3)$ for \comeng{low and high resolution GANs, respectively}.
\com{Quantitative experiments for high resolution GANs included 10,000 images in the test set.}

\subsection{DFI verification using the synthetic dataset} \label{subsection_dfi_synthetic}

\begin{figure*}[t!] %%Fig 3 in pdf
  \centering
    \includegraphics[width=0.9\textwidth]{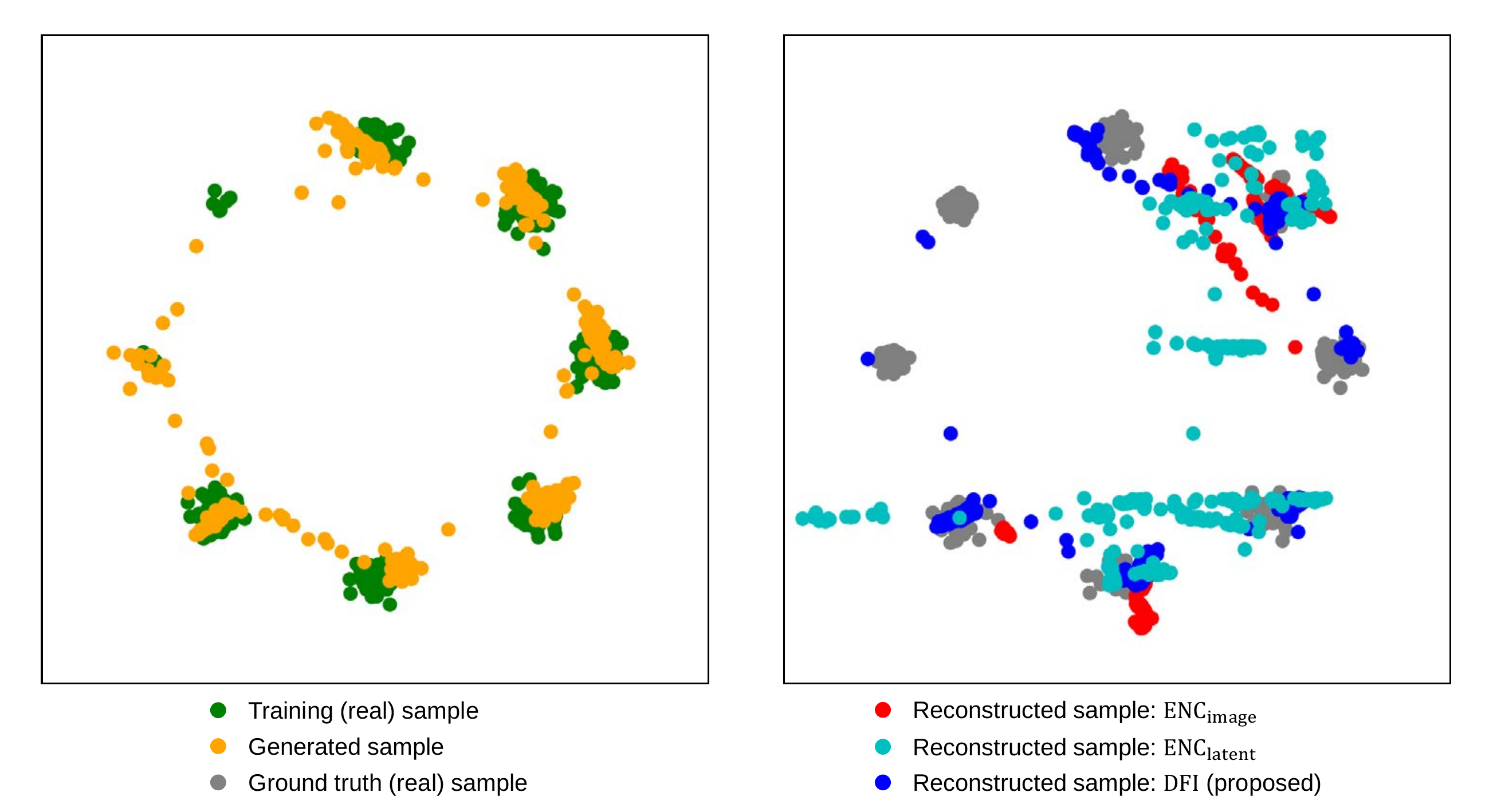}
    \caption{Inference algorithm \comeng{performances} using a synthetic dataset with eight Gaussian spreads: (left) green dots are training (real) samples, orange dots are generated samples from baseline GAN generators; (right) gray dots are ground truth (real) test samples, red dots are reconstructed samples by the \comrevision{$\mathrm{ENC_{image}}$}, the cyan dots are reconstructed samples by the \comrevision{$\mathrm{ENC^{}_{latent}}$}, and the blue dots are reconstructed samples by the proposed \comrevision{$\mathrm{DFI}$}.}
    \label{figure_synthetic_dataset}
    % \vskip -0.2in
\end{figure*}

Figure~\ref{figure_synthetic_dataset} \comeng{(left) compares} performance for the acyclic inference algorithms using the synthetic dataset. The dataset \comeng{consisted} of eight Gaussian spreads with standard deviation $= 0.1$. We \comeng{reduced} the number of samples from two Gaussian spreads at the second quadrant to induce \comeng{minor data distribution modes}, \comeng{and then} trained the GANs using real samples (green dots). The generator and discriminator \comeng{included} three FC layers with batch normalization. \comeng{Subsequently}, we \comeng{obtained} generated samples (orange dots) \comeng{by randomly producing samples using the generator.} The distributions confirm that GAN training \comeng{was} successful, \comeng{with} generated samples \comeng{covering all data distribution modes}.

Although the pre-trained GANs \comeng{covered} all modes, two modes on the second quadrant \comeng{were} rarely reproducible. This commonly incurs in GAN training, \comeng{leading to poor} diversity in sample generation. Using this pre-trained GANs, we trained \comrevision{(1) $\mathrm{ENC_{image}}$, (2) $\mathrm{ENC_{latent}}$ (the degenerated version of the proposed algorithm), and (3) $\mathrm{DFI}$ (the proposed algorithm)}. Hyper-parameters and network architecture \comeng{were identical for} all models, i.e., $\mathrm{DFI}$ \comeng{included} the discriminator (two FC layers without the final FC layer) and the connection network (two FC layers), whereas \comrevision{the encoders ($\mathrm{ENC_{image}}$ and $\mathrm{ENC_{latent}}$)} \comeng{included} four FC layers with the same architecture and model parameters as $\mathrm{DFI}$. \comeng{Each inference algorithm calculated} corresponding latent vectors \comeng{from the test samples (gray dots)}, and then regenerating \comeng{the test samples} from the latent vectors. For sufficient training, we extract the results after 50K iterations.

Figure~\ref{figure_synthetic_dataset} \comeng{(right) compares} performance \comeng{for} the inference algorithms with sample reconstruction results. The \comrevision{$\mathrm{ENC_{image}}$} (the red dots) tends to recover the right side of test samples but is incapable of recovering samples on the left side, and only five modes were recovered in this experiment; \comeng{whereas} \comrevision{$\mathrm{ENC_{latent}}$} (cyan and blue dots) \comeng{recover} many more modes after reconstruction.
This visual comparison clearly \comeng{demonstrates} the \comrevision{$\mathrm{ENC_{image}}$} \comeng{drawbacks}.

For inference algorithms with the same latent reconstruction loss, $\mathrm{DFI}$ significantly outperforms the algorithm using the \comrevision{$\mathrm{ENC_{latent}}$}. In particular, the reconstructed samples using the \comrevision{$\mathrm{ENC_{latent}}$} are inaccurate in terms of reconstruction accuracy because considerable portions of reconstructed samples (e.g. cyan dots in the middle) are far from all eight Gaussian spreads. DFI reconstructed samples are much closer to the original Gaussian spreads, i.e., more accurate results.

\comeng{Thus,} latent reconstruction loss \comeng{was} more effective than image reconstruction loss to derive accurate acyclic inference algorithms. Utilizing the pre-trained discriminator as a feature extractor \comeng{also helped to further increase} inference mapping accuracy.
\comeng{Therefore, the proposed approach to employ} latent reconstruction loss with the discriminator as a feature extractor is an effective and efficient solution for inference algorithms.

\subsection{Comparison with acyclic inference models} \label{Compare_acyclic}

\begin{figure*}[t!]
  \centering
        \centering
        \includegraphics[width=.69\textwidth]{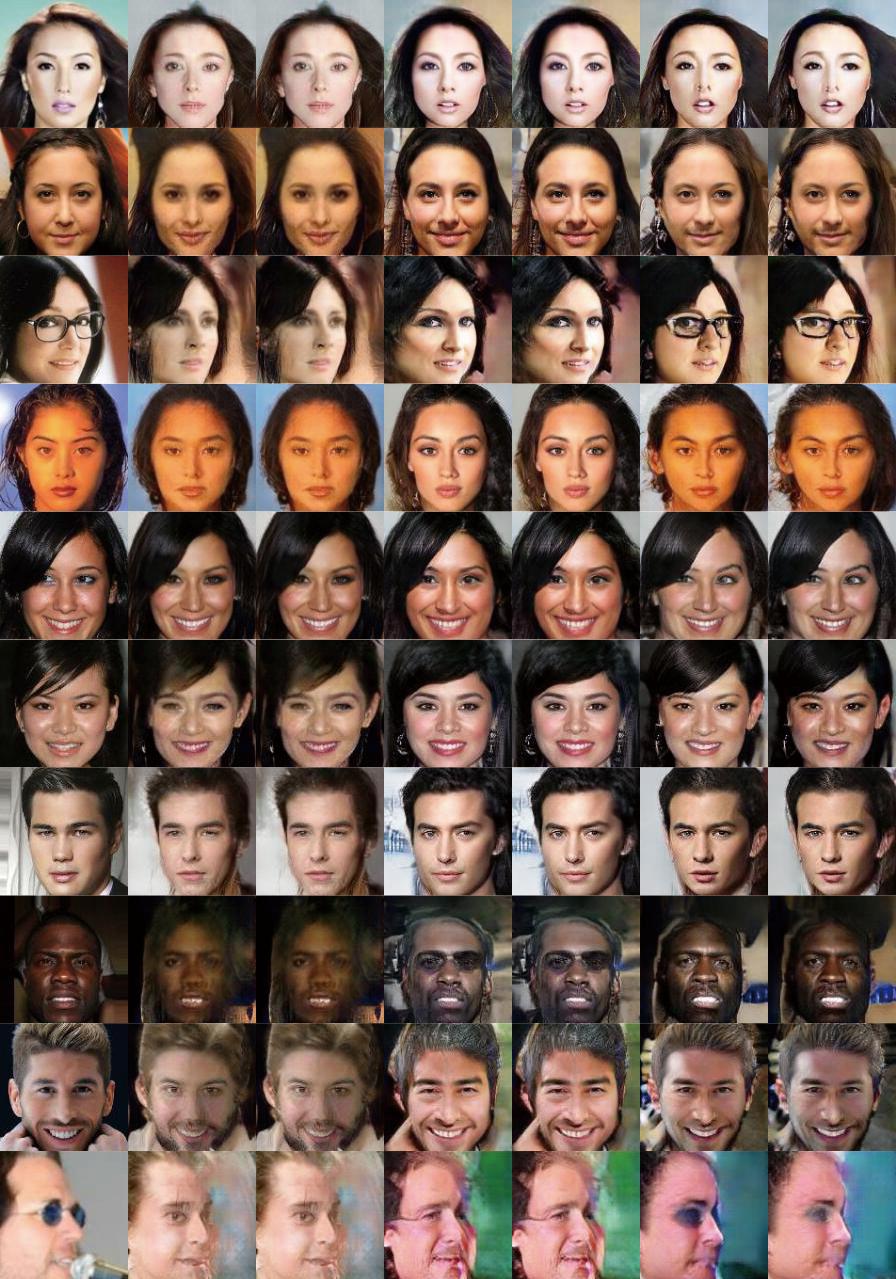}

    \small
    \vskip -0.03in
    \vskip 0.05in 
    \normalsize
    {\resizebox{0.75\textwidth}{!}
    {\begin{tabular}{c|cc|cccc}
        \toprule
                    & \multicolumn{2}{c|}{Image reconstruction loss}     & \multicolumn{4}{c}{Latent reconstruction loss}     \\
                    & $\mathrm{ENC_{image}}$ &  $\mathrm{ENC^{opt}_{image}}$ (iGAN)  & 
                    $\mathrm{ENC^{}_{latent}}$ & $\mathrm{ENC^{opt}_{latent}}$ &  $\mathrm{DFI}$ (proposed) &  $\mathrm{DFI^{opt}}$   \\
        \midrule
        PSNR (dB)    &
        {19.15} & {19.43} & 
        {13.57} & {14.15} & {14.77} & {15.66}  \\
        SSIM    &
        {0.5440} & {0.5519} & 
        {0.4271} & {0.4353} & {0.4841} & {0.4956}  \\
        \midrule
        \makecell{LPIPS}    &
        {0.1968} & {0.1949} & 
        {0.2279} & {0.2228} & \textbf{0.1931} & \textbf{0.1848}  \\
        \makecell{FAC (\%)}   &
        {87.92} & {88.04} & 
        {87.67} & {87.87} & {\textbf{89.31}} & {\textbf{89.55}} \\
        FID    &
        {23.38} & {22.68} & 
        {12.90} & {12.22} & \textbf{8.91} & \textbf{8.12}  \\
        User study(\%)   &
        {6.83} & {8.33} & 
        {6.33} & {4.17} & {\textbf{27.83}} & {\textbf{46.50}} \\
        \bottomrule
    \end{tabular}}
    }
    \vskip 0.1in
    \caption{Qualitative and quantitative comparison of various inference algorithms. \comeng{Column 1 includes} target (real) images \comeng{and the remaining columns} include reconstructed images by each method \comeng{in the order shown in the table}. All images \comeng{were} computed after 40K training steps. The table summarizes quantitative metrics. \com{We used \textit{AlexNet(lin)}~\citep{zhang2018unreasonable} environment for LPIPS perceptual loss. FAC indicates face attribute classification accuracy using 13 attributes in CelebA dataset. Smaller LPIPS and FID indicate more accurate and realistic results.} \comeng{User study participants selected} the image most similar to the target among six reconstructed images. \comeng{The number of votes in percentage is reported in the table.}}
    \label{figure_igan_compare}
    %\vskip -0.2in
\end{figure*}

\begin{figure*}[t!]
    \centering
    \includegraphics[width=.85\textwidth]{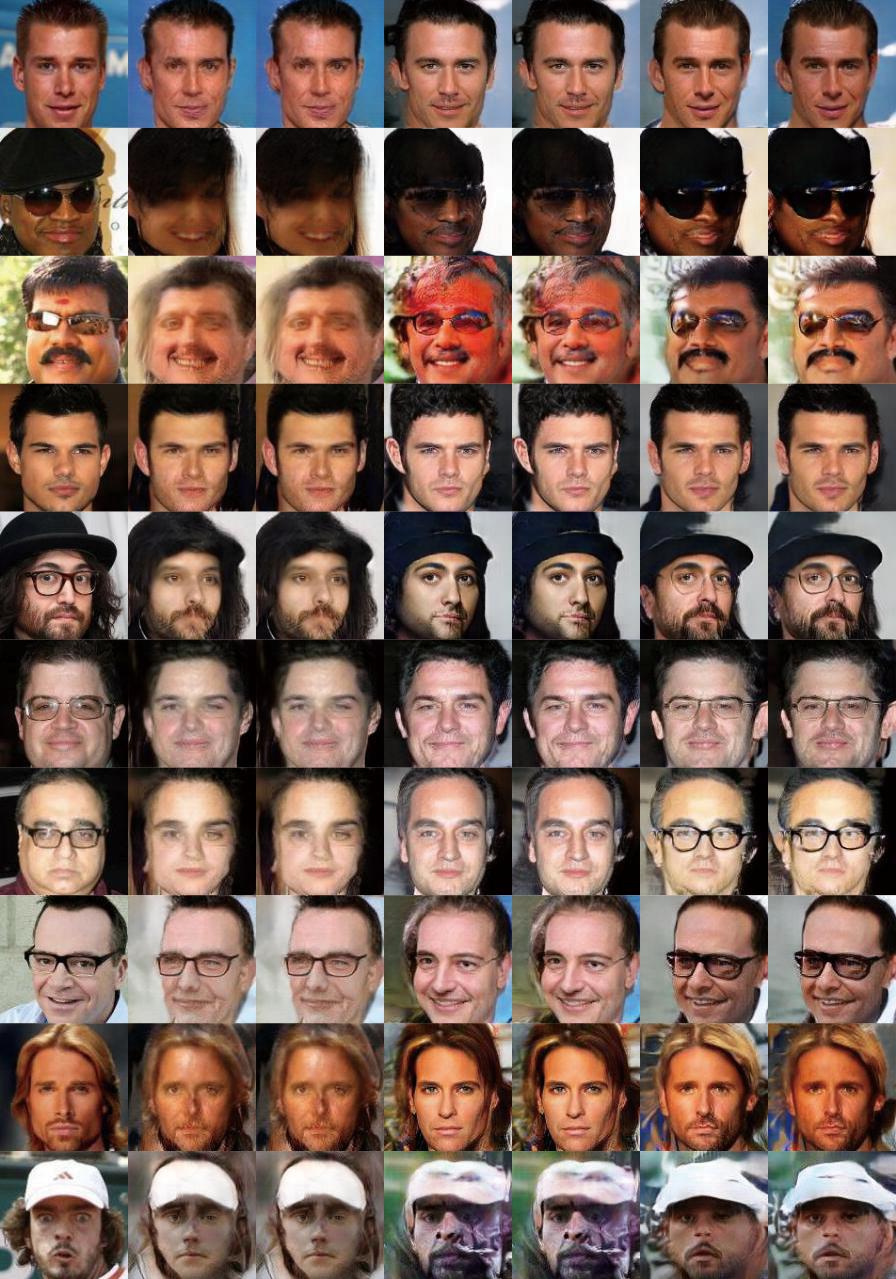}
    \vskip 0.1in
    \caption{Qualitative comparisons for various inference algorithms. Column 1 includes target (real) images and remaining columns include reconstructed images by each method in the order of Figure~\ref{figure_igan_compare}. All images were computed after 40K training steps.}
    \label{figure_igan_compare_additional}
\end{figure*}

\com{In Figure~\ref{figure_igan_compare}, we use various objective metrics for quantitatively evaluating the inference algorithms. Specifically, PSNR, SSIM, LPIPS, face attribute classification (FAC) accuracy, FID and user study results are reported for comparing DFI with the other acyclic models. For the FAC accuracy, we utilize the same classifier as STGAN~\citep{liu2019stgan}, that uses 13 attributes in CelebA dataset to measure accuracy. For the experimental results in CelebA, LPIPS exhibits similar tendency to FAC accuracy. Therefore, we choose LPIPS to assess inference algorithm semantic similarity for the remaining experiments since if can measure semantic fidelity on various datasets.}

\com{LPIPS, FAC accuracy, FID and the user study scores indicate DFI based models to be significantly superior. Although PSNR and SSIM scores from methods using image reconstruction loss are significantly higher than for DFI models, significant gaps in the user study confirm that PSNR and SSIM are not reliable metrics for this application. Inference algorithms with image reconstruction loss are expected to have higher PSNR and SSIM scores, simply because their objectives, i.e., minimizing pixel-level difference exactly match the metrics.} 

\comrevision{$\mathrm{ENC^{}_{latent}}$ and $\mathrm{ENC^{opt}_{latent}}$} results do not provide accurate fidelity (lower LPIPS). The \comrevision{$\mathrm{ENC^{}_{latent}}$} utilizes only fake samples for training the feature extractor, i.e., convolutional layers, whereas DFI exploits the discriminator feature extractor,which was trained with real and fake samples. Thus, the \comrevision{$\mathrm{ENC^{}_{latent}}$} model is incapable of capturing a common feature to represent real and fake images. Consequently, reconstruction fidelity is significantly degraded. On the other hand, their image quality, i.e., realistic and sharp, exceeds other methods using image reconstruction loss, because the inference algorithm learns to reduce image level distance regardless of the  image manifold. Consequently, it tends to produce blurry images without distinct attributes, leading to quality degradation. In contrast, inference algorithms with latent reconstruction loss generally provide high quality images after inference mapping. Thus, latent distance is more favorable to retain samples onto the image manifold, helping to improve image quality.

All \com{LPIPS,} FID assessments, and user study scores confirm that \comrevision{$\mathrm{DFI}$ and $\mathrm{DFI^{opt}}$} outperform the other models. Other inference mappings are \comeng{particularly} degraded when the input images include distinctive attributes, such as eyeglasses or a mustache; \comeng{whereas} the proposed DFI inference mapping consistently performs well, \comeng{increasing} the performance gap between \comeng{the proposed DFI mapping and others approaches} for samples with distinctive attributes. Therefore, the proposed inference mapping \comeng{was} effective in restoring semantic attributes and reconstruction results \comeng{were} semantically more accurate than other \comeng{inference mappings}.

\com{ \comeng{
Figure~\ref{figure_igan_compare_additional} compares the proposed DFI method with \comrevision{(1) encoder mapping ($\mathrm{ENC_{image}}$ and $\mathrm{ENC^{}_{latent}}$), (2) hybrid inference as suggested by iGAN \citep{ref33} ($\mathrm{ENC^{opt}_{image}}$ and $\mathrm{ENC^{opt}_{latent}}$), and (3) $\mathrm{DFI^{opt}}$}. To investigate the effect of latent reconstruction loss, we modified the encoder objective function in (1) and (2) from image reconstruction loss to latent reconstruction loss. 
}}

Reconstruction results using image reconstruction loss (\comeng{Columns 2 and 3} from Figures~\ref{figure_igan_compare} and~\ref{figure_igan_compare_additional}) are generally blurred or have missing attributes, e.g. eyeglasses, mustache, gender, wrinkles, etc., compared with \comeng{DFI reconstruction results}. These results support our argument in Section \ref{sec3_1}: latent reconstruction loss \comeng{provides} more accurate inference mapping than image reconstruction loss. \comeng{Previous iGAN studies} have shown that additional latent optimization after inference mapping (in both \comrevision{$\mathrm{ENC^{opt}_{image}}$ and $\mathrm{DFI^{opt}}$}) \comeng{effectively improves} inference accuracy. The current study found that optimization was useful to better restore the original color distribution, based on feedback from the user study. 

However, although the additional optimization fine tunes the inference mapping, it still has \comeng{computational efficiency limitations}. Therefore, \comeng{we chose DFI} without additional optimization \comeng{for subsequent} experiments \comeng{to trade-off between accuracy and computational efficiency}. 

The last row in Figures~\ref{figure_igan_compare} and \ref{figure_igan_compare_additional} \comeng{present examples where} all inference methods \comeng{performed} poorly. These poor results \comeng{were due to baseline GAN performance limitations} rather than the inference algorithms. \comeng{However,} despite the inaccurate reconstruction, \comeng{the proposed DFI approach recovered many} original semantic attributes, e.g. glasses on the right side and mustache on the left.

\subsection{Comparison with cyclic inference models}\label{Compare_cyclic}

\begin{figure*}[t!]
  \centering
    \includegraphics[width=1.0\textwidth]{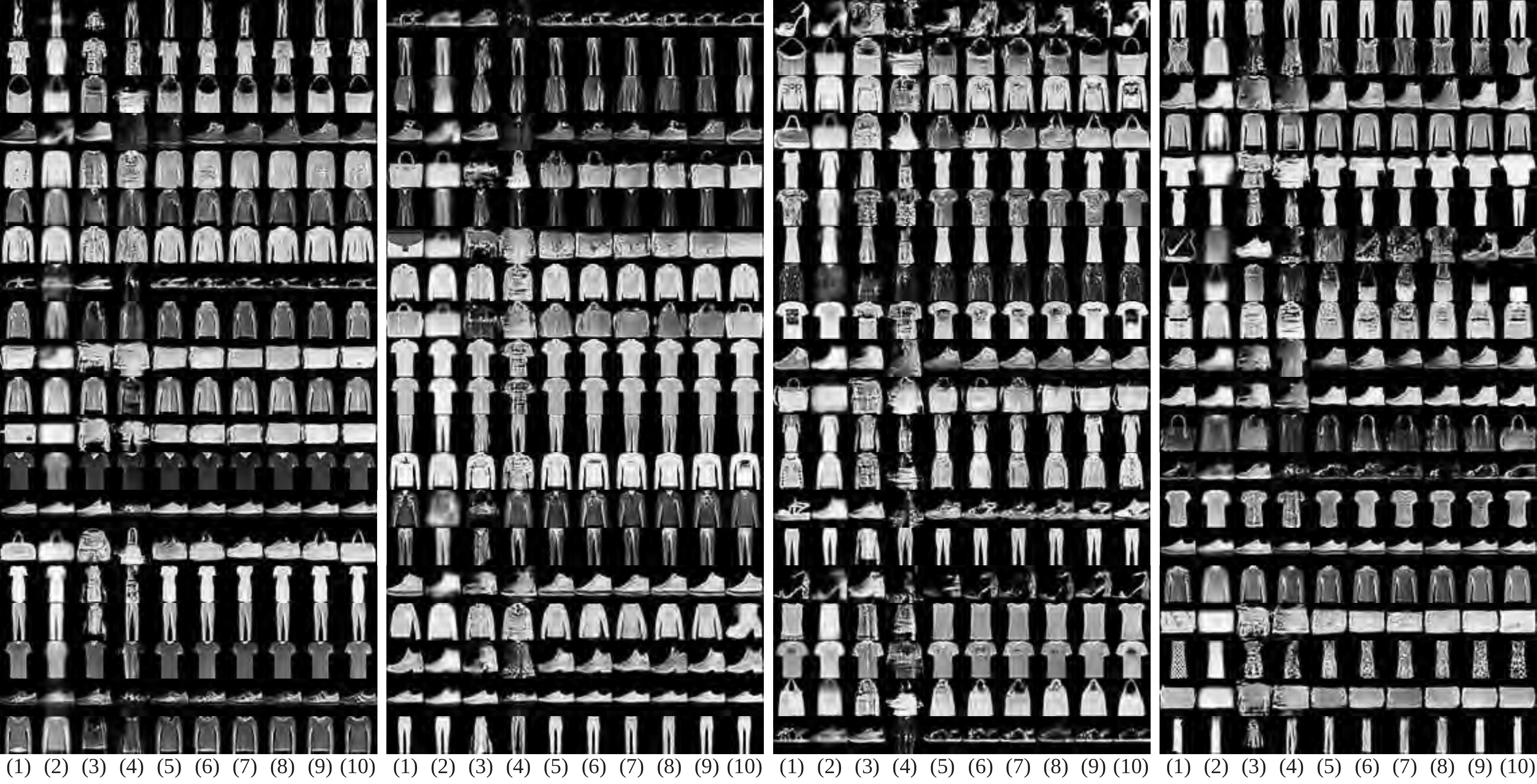}
    \vskip 0.4in
    \includegraphics[width=1.0\textwidth]{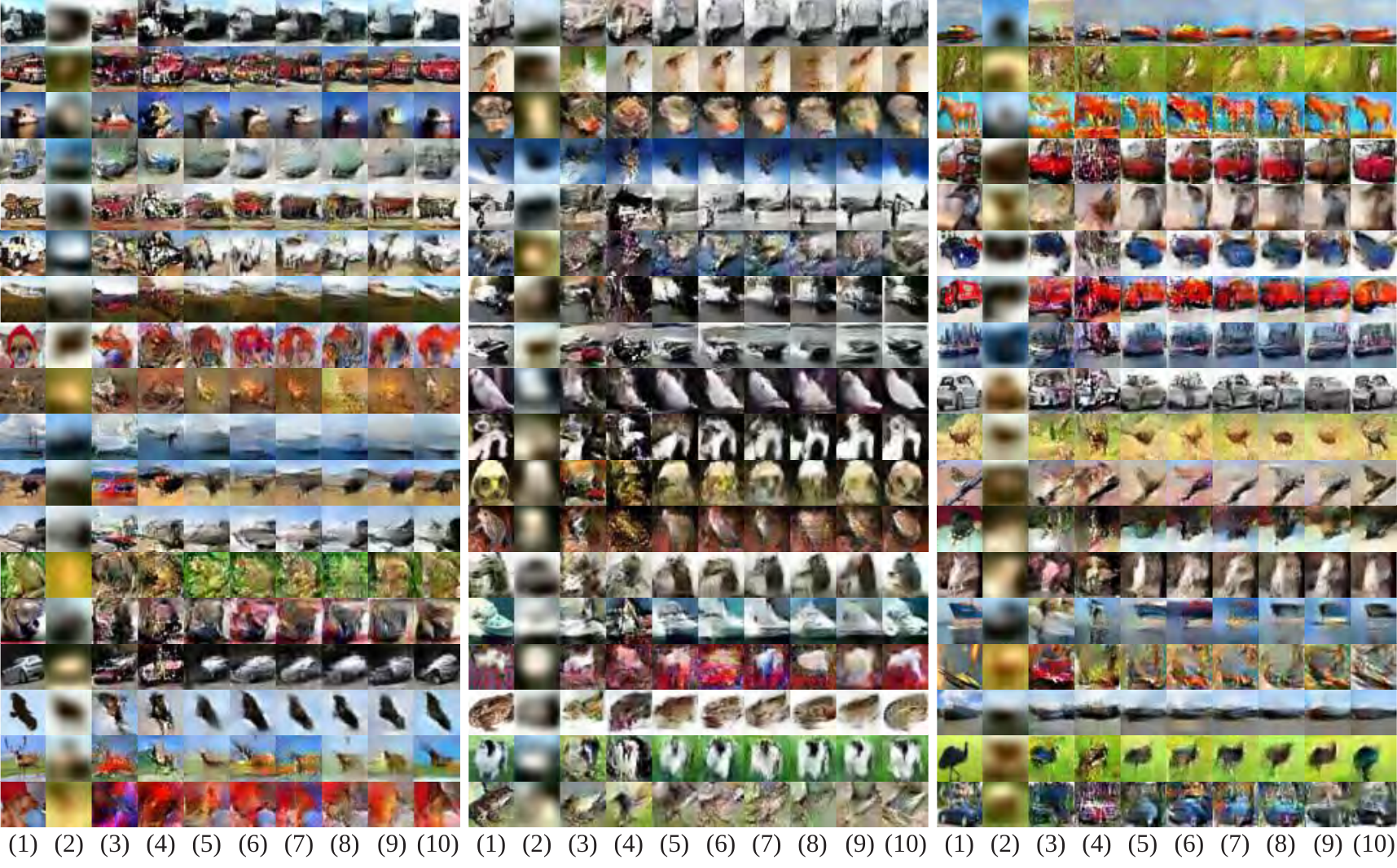}
    \vskip 0.2in
    \caption{Qualitative comparison with cyclic inference algorithms and DFI variants using FashionMNIST and CIFAR-10 datasets. Column (1) includes target (real) images, and the remainder include reconstructed images by (2) VAE, (3) ALI/BiGAN, (4) ALICE, DFI with $\{$(5) DCGAN, (6) LSGAN, (7) DFM, (8) RFGAN, (9) SNGAN, and (10) WGAN-GP$\}$.}
    \label{figure_recon_compare_appendix_fashion_cifar}
\end{figure*}

\begin{figure*}[t!]
  \centering
    \includegraphics[width=1.0\textwidth]{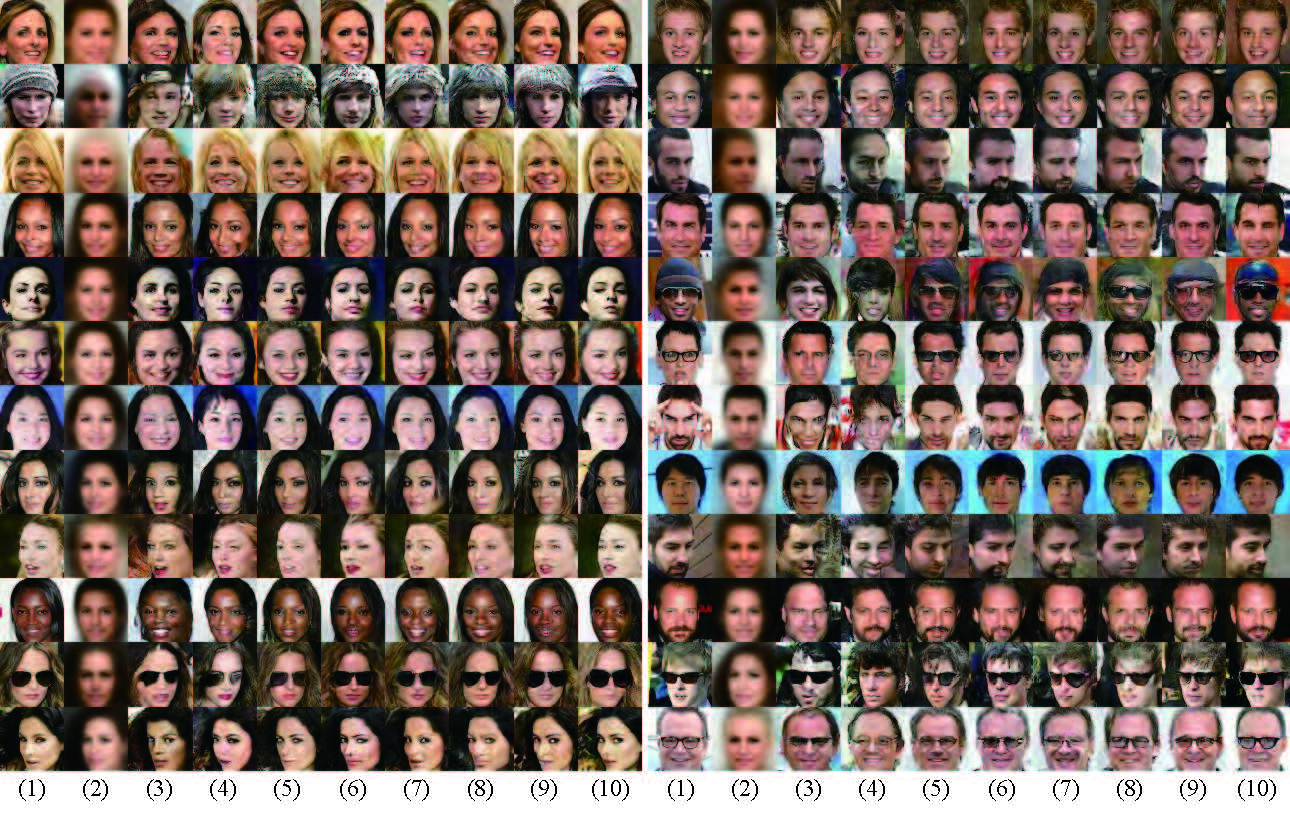}
    % \vskip -0.15in
    \caption{Qualitative comparison with cyclic inference algorithms and DFI variants using the CelebA dataset. Column (1) includes target (real) images and the remainder include reconstructed images by (2) VAE, (3) ALI/BiGAN, (4) ALICE, DFI with $\{$(5) DCGAN, (6) LSGAN, (7) DFM, (8) RFGAN, (9) SNGAN, and (10) WGAN-GP$\}$.}
    \label{figure_recon_compare_appendix_celeba}
\end{figure*}

\begin{table*}[t!]
  \centering
  \small
  \resizebox{\textwidth}{!}
  {
    \begin{tabular}{c c | c c c | c c c c c c}
        \hline
            \multicolumn{2}{c|}{} &  \multicolumn{3}{c|}{Cyclic models} & \multicolumn{6}{c}{ DFI with various GAN models} \\
            Dataset & Metric & VAE & ALI/BiGAN & ALICE & DCGAN & LSGAN & DFM & RFGAN & SNGAN & WGAN-GP \\
            \hline
            \multirow{3}{*}{\makecell{Fashion \\ MNIST}} &
            {LPIPS} &
            {\makecell{0.1287 \\ $\pm$ 0.0017}} & 
            {\makecell{0.0869 \\ $\pm$ 0.0097}} & 
            {\makecell{0.0898 \\ $\pm$ 0.0177}} & 
            {\makecell{0.0351 \\ $\pm$ 0.0014}} & 
            {\makecell{\textbf{0.0269} \\ $\pm$ 0.0009}} & 
            {\makecell{0.0352 \\ $\pm$ 0.0013}} & 
            {\makecell{0.0345 \\ $\pm$ 0.0017}} & 
            {\makecell{\textbf{0.0306} \\ $\pm$ 0.0004}} & 
            {\makecell{0.0366 \\ $\pm$ 0.0012}} \\  \cline{2-11} & 
             
            {FID} &
            {\makecell{90.94 \\ $\pm$ 2.78}} & 
            {\makecell{69.99 \\ $\pm$ 23.73}} & 
            {\makecell{62.10 \\ $\pm$ 13.77}} & 
            {\makecell{25.73 \\ $\pm$ 1.27}} & 
            {\makecell{19.76 \\ $\pm$ 0.60}} & 
            {\makecell{25.88 \\ $\pm$ 1.00}} & 
            {\makecell{25.26 \\ $\pm$ 3.06}} & 
            {\makecell{\textbf{15.53} \\ $\pm$ 1.00}} & 
            {\makecell{\textbf{11.85} \\ $\pm$ 0.28}} \\

            \hline
            \multirow{3}{*}{CIFAR10} &
            {LPIPS} &
            {\makecell{0.4839 \\ $\pm$ 0.0014}} & 
            {\makecell{0.1094 \\ $\pm$ 0.0081}} & 
            {\makecell{0.1108 \\ $\pm$ 0.0061}} & 
            {\makecell{0.0744 \\ $\pm$ 0.0033}} & 
            {\makecell{\textbf{0.0716} \\ $\pm$ 0.0032}} & 
            {\makecell{0.0745 \\ $\pm$ 0.0022}} & 
            {\makecell{\textbf{0.0735} \\ $\pm$ 0.0021}} & 
            {\makecell{0.2799 \\ $\pm$ 0.0164}} & 
            {\makecell{0.0758 \\ $\pm$ 0.0019}} \\  \cline{2-11} &
            
            {FID} &
            {\makecell{234.03 \\ $\pm$ 1.30}} & 
            {\makecell{54.04 \\ $\pm$ 5.66}} & 
            {\makecell{56.43 \\ $\pm$ 8.44}} & 
            {\makecell{44.83 \\ $\pm$ 1.92}} & 
            {\makecell{\textbf{39.67} \\ $\pm$ 1.62}}  & 
            {\makecell{46.23 \\ $\pm$ 2.83}} & 
            {\makecell{46.79 \\ $\pm$ 5.97}} & 
            {\makecell{61.88 \\ $\pm$ 7.75}} & 
            {\makecell{\textbf{31.54} \\ $\pm$ 1.49}} \\
            \hline
            
            \multirow{3}{*}{CelebA} &
            {LPIPS} &
            {\makecell{0.3535 \\ $\pm$ 0.0026}} & 
            {\makecell{0.1573 \\ $\pm$ 0.0213}} & 
            {\makecell{0.1424 \\ $\pm$ 0.0052}} & 
            {\makecell{\textbf{0.0980} \\ $\pm$ 0.0012}} & 
            {\makecell{0.1237 \\ $\pm$ 0.0185}} & 
            {\makecell{\textbf{0.0980} \\ $\pm$ 0.0005}} & 
            {\makecell{0.0994 \\ $\pm$ 0.0009}} & 
            {\makecell{0.0997 \\ $\pm$ 0.0087}} & 
            {\makecell{0.1033 \\ $\pm$ 0.0011}} \\  \cline{2-11} &
             
            {FID} &
            {\makecell{94.96 \\ $\pm$ 0.52}} & 
            {\makecell{19.56 \\ $\pm$ 5.61}} & 
            {\makecell{16.36 \\ $\pm$ 1.92}} & 
            {\makecell{12.86 \\ $\pm$ 0.15}} & 
            {\makecell{24.99 \\ $\pm$ 14.17}} & 
            {\makecell{13.24 \\ $\pm$ 0.16}} & 
            {\makecell{14.02 \\ $\pm$ 1.08}} & 
            {\makecell{\textbf{11.92} \\ $\pm$ 3.44}} & 
            {\makecell{\textbf{8.73} \\ $\pm$ 0.78}} \\
        \hline
    \end{tabular}
    }
    \normalsize
    \vskip 0.1in
    \caption{ \com{Quantitative comparisons with cyclic inference algorithms and DFI variants for Fashion MNIST, CIFAR10, and CelebA datasets using LPIPS and FID. Bold values indicate Top-2 scores for each metric and dataset.}}
    % \vskip -0.1in
    \label{table_recon_compare}
    %\vskip -0.2in
\end{table*}

\begin{table}[t!]
  \centering
  \small
  \resizebox{0.98\columnwidth}{!}{
    \begin{tabular}{c | c c c}
        \hline
            Metric & DCGAN-0GP & SNGAN & SNGAN-0GP \\
        \hline
            LPIPS &
            {\makecell{0.2147 \\ $\pm$ 0.0014}} &
            {\makecell{0.2116 \\ $\pm$ 0.0002}} &
            {\makecell{\textbf{0.1931} \\ $\pm$ 0.0015}} \\ 
        \hline
            FID &
            {\makecell{15.10 \\ $\pm$ 0.54}} &
            {\makecell{16.09 \\ $\pm$ 0.30}} &
            {\makecell{\textbf{8.91} \\ $\pm$ 0.12}} \\
        \hline
    \end{tabular}}
    \normalsize
    \vskip 0.1in
    \caption{ Quantitative comparison of cyclic inference algorithms and DFI variants for the CelebA dataset using LPIPS and FID: top score among high resolution (HR) GANs. Bold values indicate Top-1 scores for each metric.}
    % \vskip -0.1in
    \label{table_recon_compare_hr}
    %\vskip -0.2in
\end{table}

\comeng{Figures~\ref{figure_recon_compare_appendix_fashion_cifar} and \ref{figure_recon_compare_appendix_celeba} compare the proposed DFI approach with} VAE, ALI/BiGAN, and ALICE representative generative models that allow inference mapping adopting the six baseline GANs \comeng{discussed above. Table~\ref{table_recon_compare} show}s corresponding reconstruction accuracy in terms of \com{LPIPS and} FID. 

Reconstructed images from VAE are blurry and lose detailed structures because it \comeng{was} trained with image reconstruction loss. Less frequently appearing \comeng{training dataset} attributes, e.g. mustache or baldness, \comeng{were} rarely recovered due to popularity bias. ALI/BiGAN and ALICE restore sharper images than VAE, \comeng{but do not} effectively recover important \comeng{input image characteristics}, e.g. identity, and occasionally generate completely different images from the inputs.

\comeng{In contrast,} reconstructed images from DFI variants exhibit consistently better visual quality than VAE, ALI/BiGAN, and ALICE. DFI training focused on accurate inference mapping, without influencing \comeng{baseline GAN performance}. Hence, reconstructed image quality from DFI models is identical to that of the baseline unidirectional GANs: sharp and realistic. DFI variants consistently provide more accurate reconstructions, i.e., faithfully reconstruct the input images including various facial attributes; whereas VAE, ALI/BiGAN, and ALICE often fail to handle \comeng{these aspects}. Thus, the proposed algorithm accurately estimates the latent vector corresponding to the input image and retains image quality better than competitors. 

\comeng{Table~\ref{table_recon_compare} confirms that} inference accuracy \comeng{for DFI based} models significantly outperform VAE, ALI/BiGAN, and ALICE \comeng{for LPIPS and FID metrics, similar to the case for qualitative comparisons. In addition, Table~\ref{table_recon_compare_hr} supports the scalability of DFI for high resolution GANs. Unlike other cyclic inference algorithms, our DFI does not influence (degrade) the generation quality of baseline GANs and still provides the robust and consistent performance in inference mapping.}

% \begin{figure}[t!]
% \centering
%         \centering
%         \includegraphics[width=\columnwidth]{figure_dfi_vgg_compare.pdf} \\
%         \small
%         \centering
%         \vskip 0.1in
%         \resizebox{0.8\columnwidth}{!}{
%         {\begin{tabular}{c | c c}
%         \toprule
%             {} &
%             {\comrevision{$\mathrm{DFI}$}} &
%             {\comrevision{$\mathrm{DFI-VGG16}$}} \\
%         \midrule
%             {LPIPS} &
%             {\textbf{0.1931}} &
%             {0.1935} \\
%             {\comrevision{FAC (\%)}} &
%             {\comrevision{\textbf{89.31}}} &
%             {\comrevision{88.44}} \\
%             {FID} &
%             {\textbf{8.91}} &
%             {9.01} \\
%             {User study (\%)} &
%             {\textbf{78.67}} &
%             {21.33} \\
%         \bottomrule
%         \end{tabular}
%         }
%         }
%         \normalsize
%     \normalsize
    
%     \vskip 0.05in
%     \caption{Proposed DFI feature extractor effects. Column (1) includes the target (real) images, (2) includes DFI reconstructed images, and (3) includes images reconstructed from a \comrevision{$\mathrm{DFI-VGG16}$} using VGG16 as the feature extractor. Experimental setting and metrics were identical to those for Figure~\ref{figure_igan_compare}. }
%     \label{figure_dfi_vgg}
% \end{figure}

\begin{figure*}[t!]
  \centering
    \includegraphics[width=2\columnwidth]{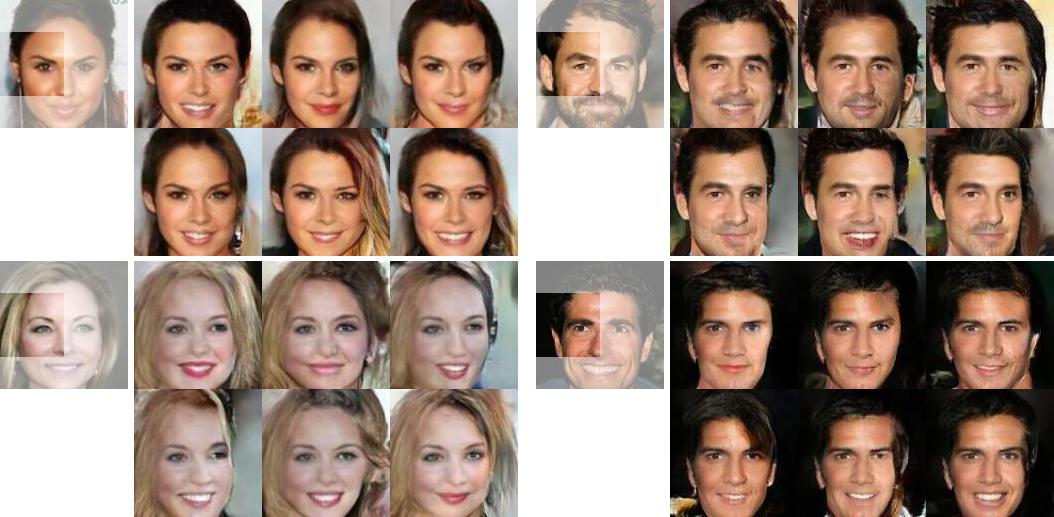}
    % \vskip -0.15in
    \caption{Spatially conditioned image generation for the CelebA dataset using the proposed SCGAN approach. Row (1) includes input images (inside box) and outer images (outside box), (2) and (3) include SCGAN generated images.
    }
    \label{figure_scgan_celeba}
\end{figure*}

\begin{figure*}[t!]
  \centering
    \includegraphics[width=2\columnwidth]{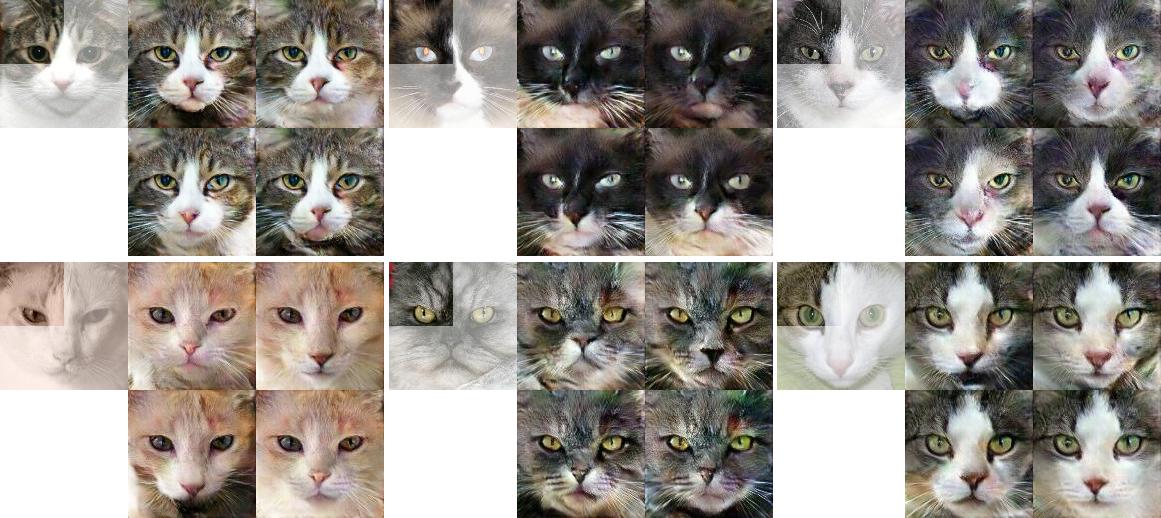}
    % \vskip -0.15in
    \caption{Spatially conditioned image generation using the proposed SCGAN approach for the cat head dataset. Row (1) includes input images (inside box) and their original outer images (outside box), (2) and (3) include SCGAN generated images.
    }
    \label{figure_scgan_cat}
\end{figure*}

\subsection{\comrevision{Ablation study on DFI}} \label{subsection_dfi_ablation}

\begin{figure}[t!]
\centering
        \centering
        \includegraphics[width=\columnwidth]{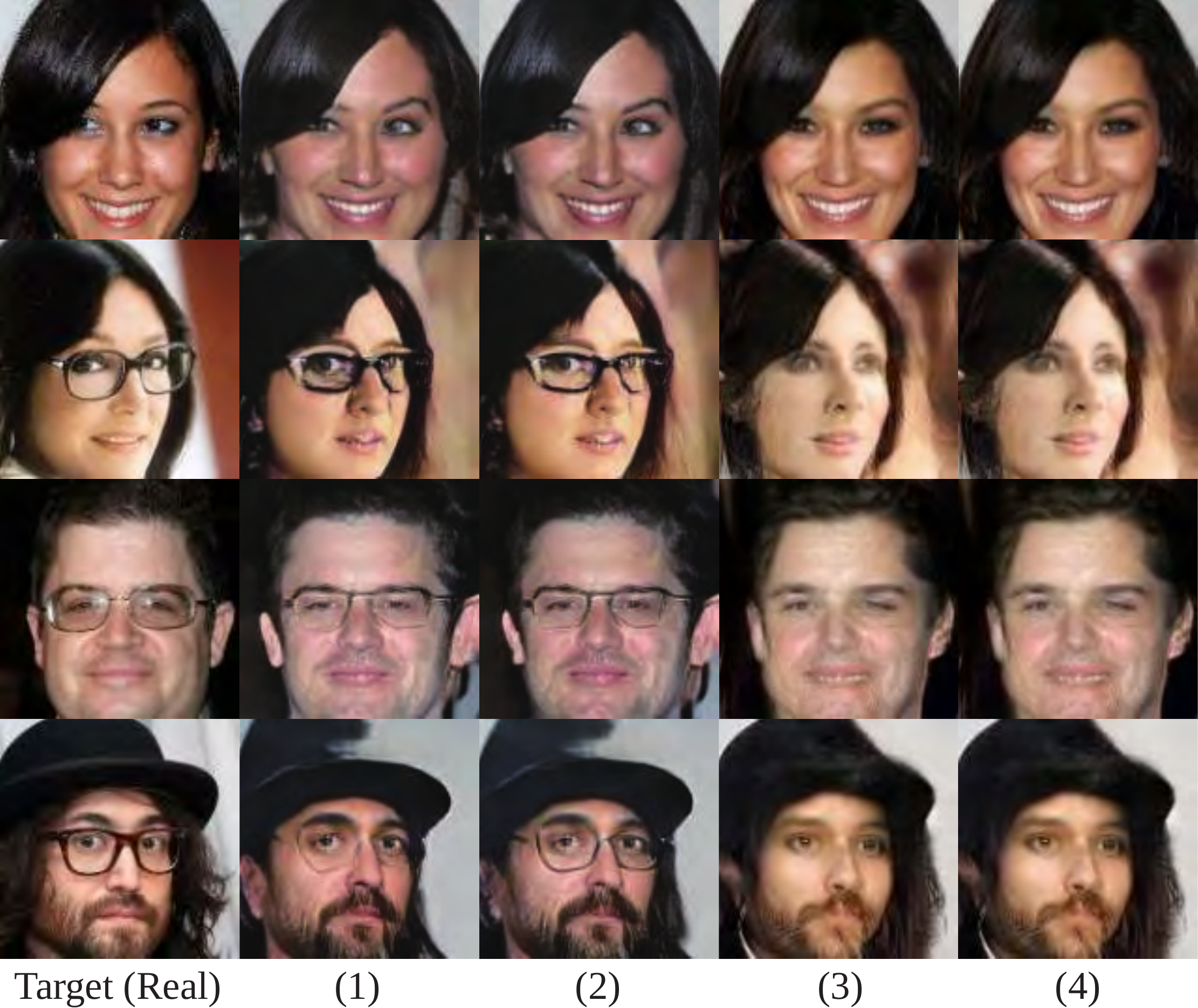} \\
        \small
        \centering
        \vskip 0.1in
        \resizebox{0.8\columnwidth}{!}{
        {\begin{tabular}{c | c c c c}
        \toprule
            {} &
            {$\mathrm{DFI}$} &
            {$\mathrm{DFI^{opt}}$} &
            {$\mathrm{DFI_{image}}$} &
            {$\mathrm{DFI^{opt}_{image}}$} \\
        
        \midrule
            {PSNR (dB)} &
            {14.77} &
            {15.66} &
            {19.23} &
            {\textbf{19.28}} \\
            
            {SSIM} &
            {0.4841} &
            {0.4956} &
            {0.5493} & 
            {\textbf{0.5501}} \\
        
        \midrule
            {LPIPS} &
            {0.1931} &
            {\textbf{0.1848}} &
            {0.1917} &
            {0.1913} \\
            
            {FAC (\%)} &
            {89.31} &
            {\textbf{89.55}} &
            {88.30} & 
            {88.31} \\
            
            {FID} &
            {8.91} &
            {\textbf{8.12}} &
            {21.33} & 
            {21.39} \\
        \bottomrule
        \end{tabular}
        }
        }
        \normalsize
    \normalsize
    
    \vskip 0.05in
    \caption{\comrevision{Ablation study on proposed DFI. The first column includes the target (real) images, (1) includes $\mathrm{DFI}$ reconstructed images, (2) includes $\mathrm{{DFI}^{opt}}$ reconstructed images, (3) includes $\mathrm{{DFI}_{image}}$ reconstructed images, and (4) includes $\mathrm{{DFI}^{opt}_{image}}$ reconstructed images, respectively. Experimental setting and metrics are identical to those for Figure~\ref{figure_igan_compare}.} }
    \label{figure_dfi_ablation}
\end{figure}

\comrevision{To understand the effect of latent reconstruction on $\mathrm{DFI}$, we conduct two experiments; (1) $\mathrm{DFI_{image}}$ and (2) $\mathrm{DFI^{opt}_{image}}$. For both experiments, the training strategy is identical to $\mathrm{DFI}$, i.e. a fixed discriminator for $\mathrm{D^f}$ and a trainable $\mathrm{CN}$ network. $\mathrm{DFI_{image}}$ utilizes the image reconstruction loss instead of the latent reconstruction loss. $\mathrm{DFI^{opt}_{image}}$ performs an  additional optimization on top of $\mathrm{DFI_{image}}$.}

\comrevision{Figure~\ref{figure_dfi_ablation} demonstrates qualitative and quantitative comparisons. Compared to the results with the latent reconstruction loss, the results from $\mathrm{DFI_{image}}$ and $\mathrm{DFI^{opt}_{image}}$ lose semantic details and quality. Even though some samples show reasonable quality, they generally lose details such as facial expressions and glasses. For example, in the fourth row in Figure~\ref{figure_dfi_ablation}, the results with the image reconstruction loss do not preserve details, whereas the results with the latent reconstruction loss do so.  In Table in Figure~\ref{figure_dfi_ablation},the LPIPS score of $\mathrm{DFI_{image}}$ is better than the proposed $\mathrm{DFI}$. However, its FID score is worse than $\mathrm{DFI}$. This is because the methods with the image reconstruction loss are optimized to reduce the pixel-level distance that leads high structural similarity regardless of its quality. Meanwhile, FID is more robust to small structural difference than LPIPS, thereby more appropriate to measure semantic similarity. This is analogous when the examples using the image reconstruction loss are compared with the examples using the latent reconstruction loss; the method using the latent reconstruction loss preserves image quality better. Similarly, despite $\mathrm{DFI_{image}}$ achieves the best  LPIPS score among all methods that do not utilize the  optimization, the image quality of $\mathrm{DFI_{image}}$ is worse than that of $\mathrm{DFI}$. Comparing $\mathrm{DFI_{image}}$ and $\mathrm{ENC_{image}}$, we observe similar visual quality and tendency. This result is consistent with our statement in Section~\ref{sec3_1} and the simulation experiment in Section~\ref{subsection_dfi_synthetic}. Because the image reconstruction loss utilizes real data for training the inference model although the generator may not be able to create them (i.e. undefined data), both $\mathrm{DFI_{image}}$ and $\mathrm{ENC_{image}}$ suffer from the inevitable errors caused by those undefined data. Despite the limitation of the image reconstruction loss, we observe that $\mathrm{DFI_{image}}$ enjoys the quantitative  improvement over $\mathrm{ENC_{image}}$ owing to the effective feature extractor (i.e. a discriminator).}

\subsection{DFI qualitative evaluation}

\begin{figure}[t!] %%Fig 11 in pdf
\centering
    \centering
    \includegraphics[width=\columnwidth]{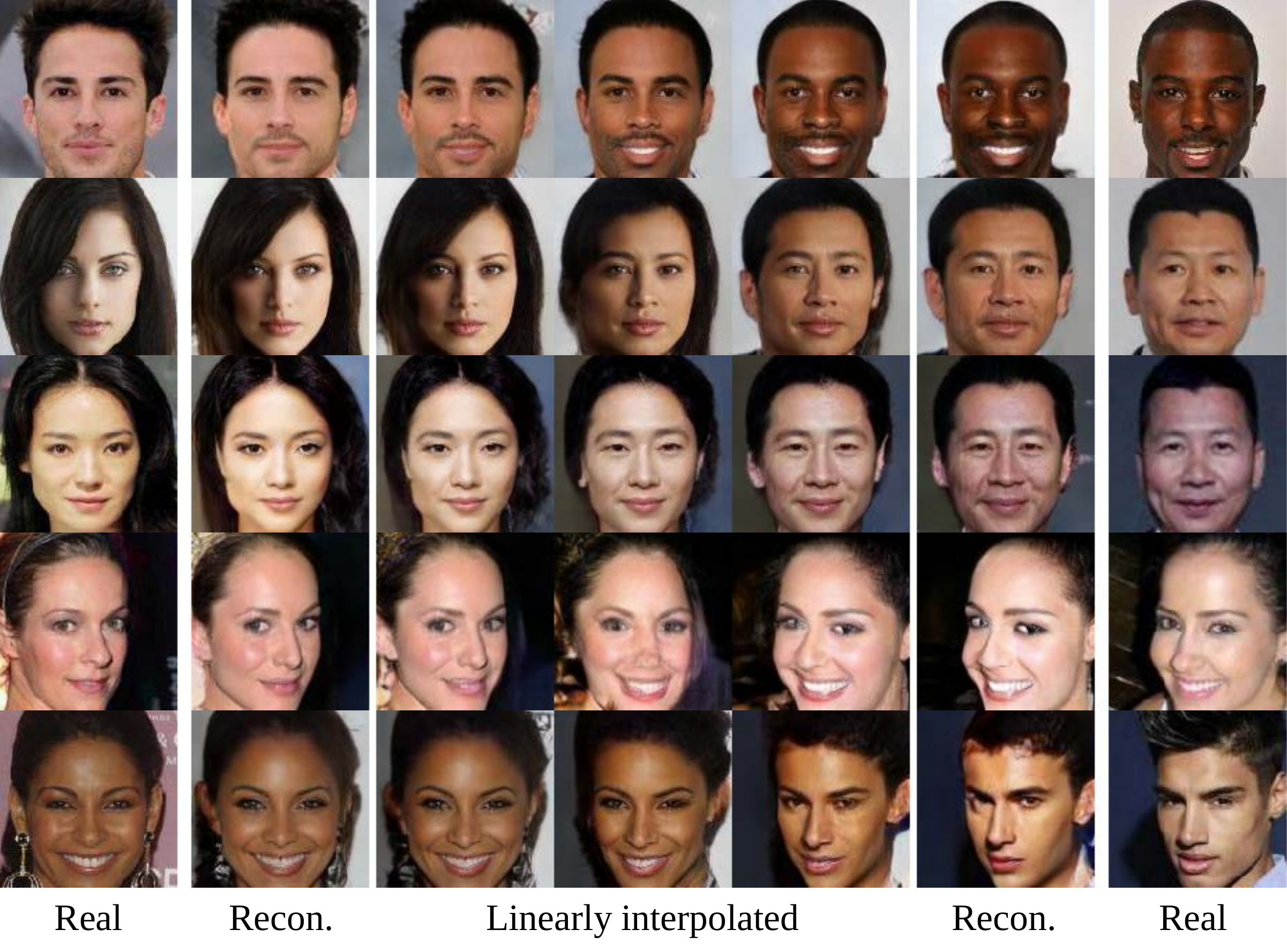} \\
    \caption{\com{Column (1) includes input images, (2)--(6) include generated images using linearly interpolated latent vector, and (7)) include latent space walking results for two inferred latent spaces using column(1) images.}}
    \label{fig_dfi_qualitative_linear_interp}
\end{figure}

\begin{figure}[t!]
\centering
    \centering
    \includegraphics[width=\columnwidth]{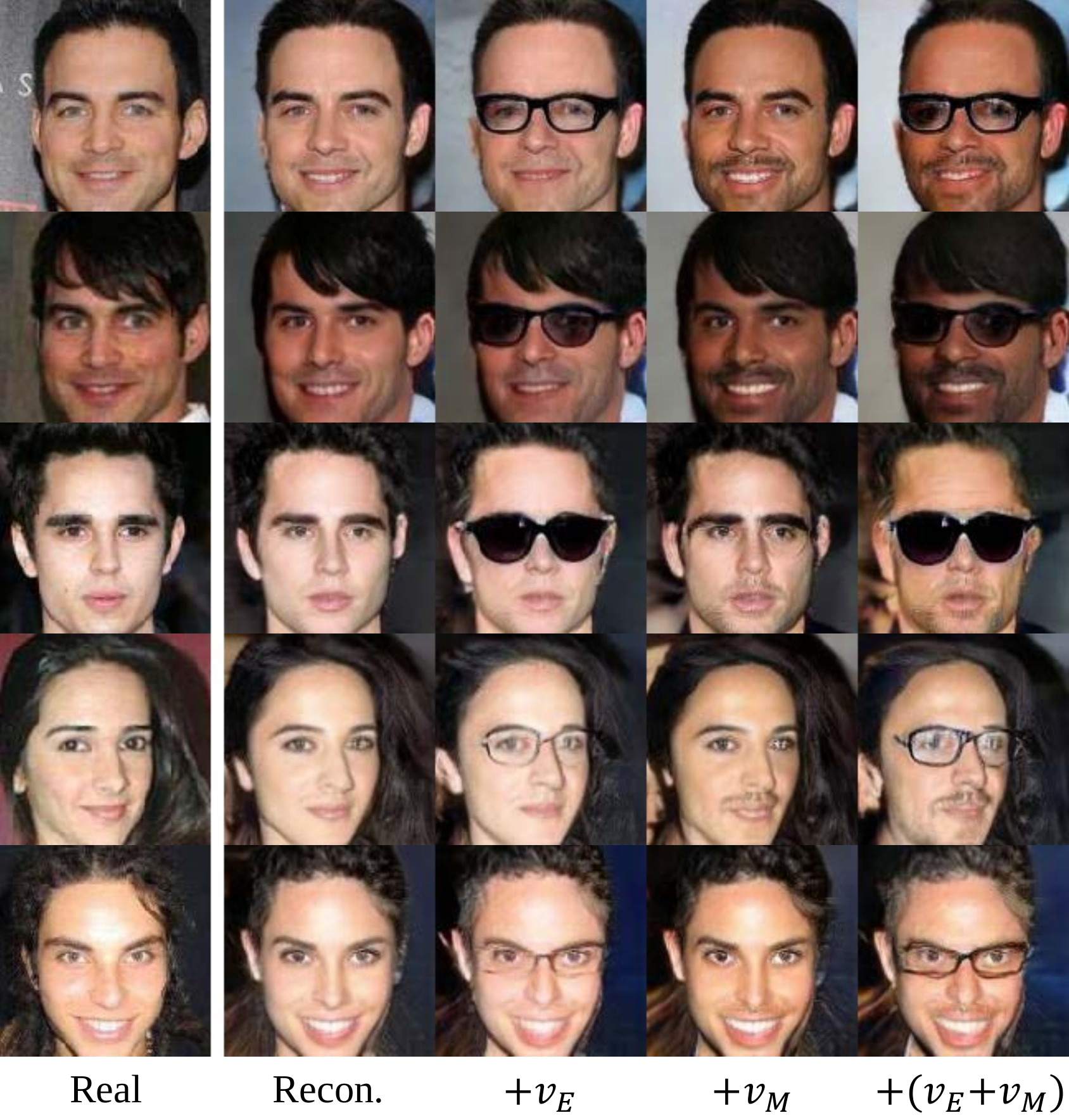} \\
    \caption{\com{Semantic image editing results using vector arithmetic on GAN latent space. Column (1) includes the original input image, (2) includes the reconstructed image using inferred latent vector by DFI, (3)--(5) include results from adding eyeglasses, mustache, and both vectors to the latent vector, respectively.}}
    \label{fig_dfi_qualitative_vector_arithmetic}
\end{figure}

\com{
To verify that DFI produced semantically accurate inference mapping, we applied latent space walking on the inferred latent vector. For two real images $x_1$ and $x_2$, we obtained inferred latent vectors $z'_1$ and $z'_2$ using DFI. Then we linearly interpolated $z'_L = \alpha z'_1 + (1 - \alpha) z'_2$, where $\alpha \in [0, 1]$. Figure~\ref{fig_dfi_qualitative_linear_interp} shows images generated using $z'_L$, where columns (2)--(6) include interpolated images for $\alpha=0.00, 0.25, 0.50, 0.75, 1.00$, respectively. If DFI incorrectly mapped the real images to the latent manifold, reconstructed images would exhibit transitions or unrealistic images. However, all reconstructed images exhibit semantically smooth transitions on the image space, e.g. skin color, hair shape, face orientation and expressions all change smoothly.}

\com{
Figure~\ref{fig_dfi_qualitative_vector_arithmetic} show vector arithmetic results for adding eyeglasses and mustache vector attributes ($v_E$ and $v_M$, respectively):
}

\com{
\begin{equation}
    {v_{E} = \frac{1}{2} (v_{EO}^{male} - v_{OO}^{male}) + \frac{1}{2} (v_{EO}^{female} - v_{OO}^{female})},
\end{equation}
}
\com{
\[v_{M} = v_{OM}^{male} - v_{OO}^{male}, \]
}

\noindent \com{where $v$ with any superscripts and subscripts are mean sample vectors inferred by DFI; $E$ and $M$ in subscripts indicate eyeglasses and mustache attributes \comeng{presence, respectively}, in sample images, and $O$ indicates \comeng{non-presence of an attribute}. We used 20 images to obtain the mean inferred vector for each group. Thus, Simple vector arithmetic on the latent vector can manipulate images, e.g. adding eyeglasses, mustache, or both.
Therefore, DFI successfully establishes semantically accurate mapping from image to latent space.}

\subsection{Feature extractor effects }\label{subsction_dfi_vgg}

\begin{figure*}[t!]
  \centering
    \includegraphics[width=0.8\linewidth]{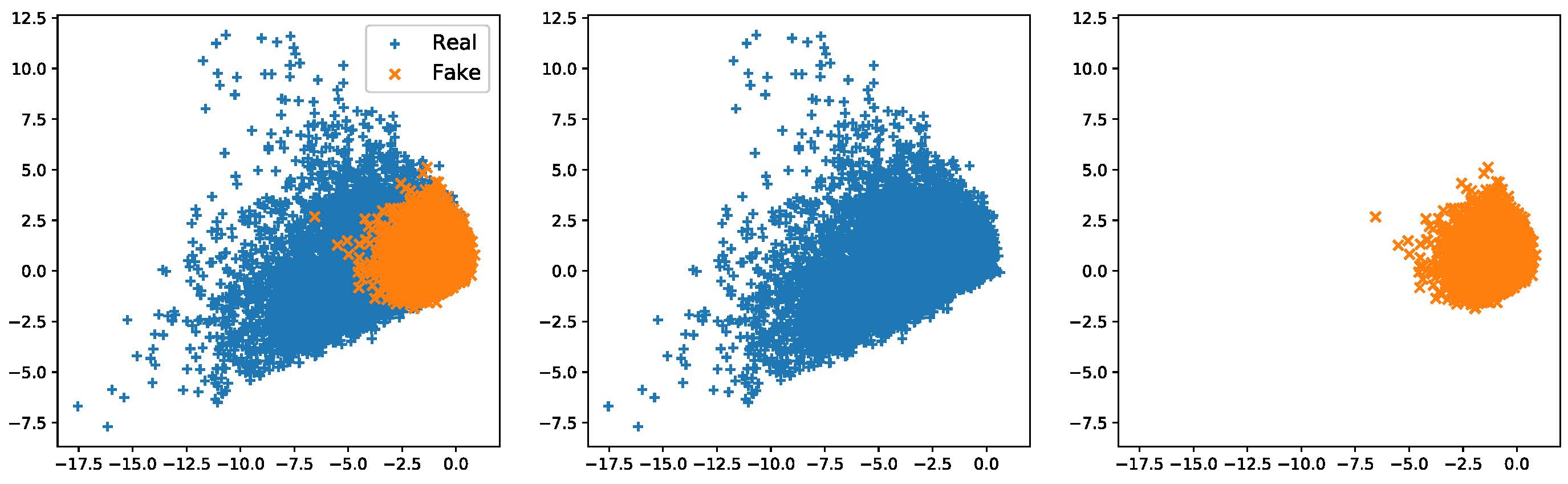}
    % \vskip -0.15in
    \caption{\com{Visualizations for ${\mathrm{D}^{\mathrm{f}}(x)}$ and $\mathrm{D}^\mathrm{f}(\mathrm{G}(z))$ using two most significant principal component axis projection. Columns (2) and (3) show real and fake samples separately, respectively, with the same axis scale as first column to more easily visualize the overlap area.}}
    \label{figure_discriminator_feature_pca}
    %\vskip -0.2in
\end{figure*}

\begin{figure}[t]
\centering
        \centering
        \includegraphics[width=\columnwidth]{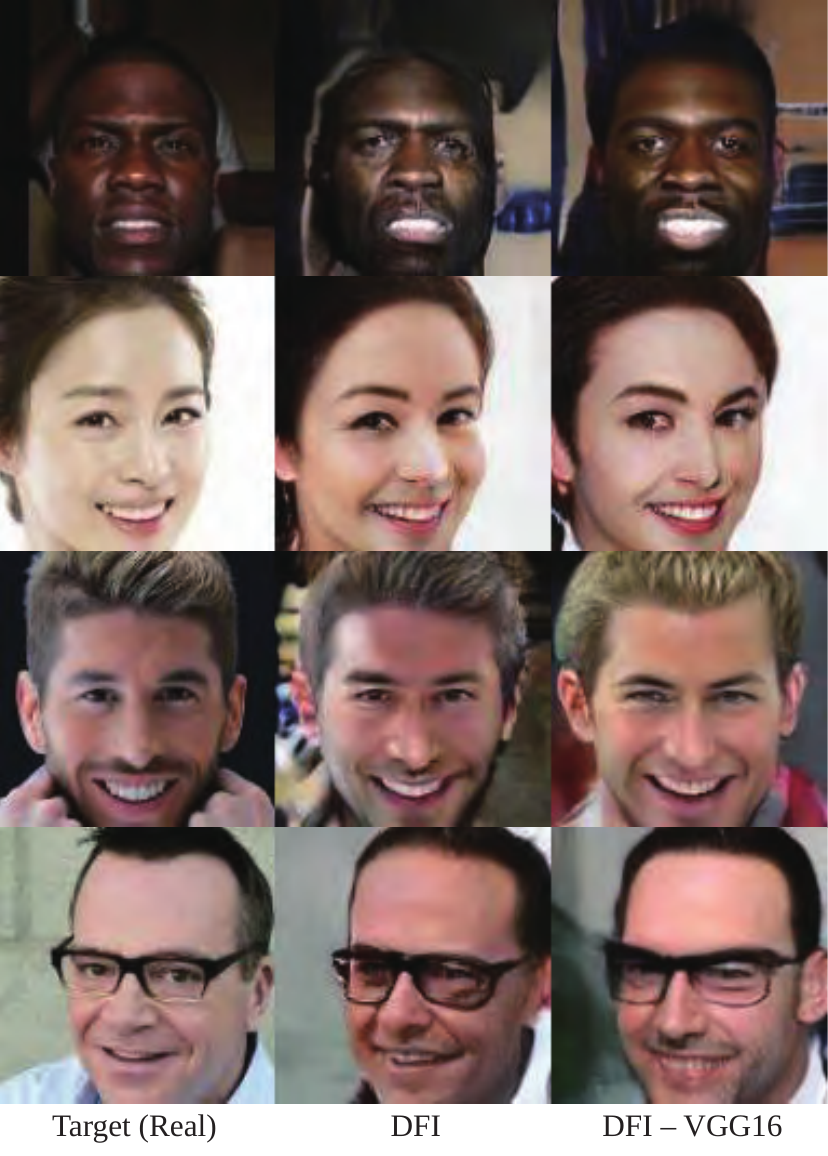} \\
        \small
        \centering
        \vskip 0.1in
        \resizebox{0.8\columnwidth}{!}{
        {\begin{tabular}{c | c c}
        \toprule
            {} &
            {\comrevision{$\mathrm{DFI}$}} &
            {\comrevision{$\mathrm{DFI-VGG16}$}} \\
        \midrule
            {LPIPS} &
            {\textbf{0.1931}} &
            {0.1935} \\
            {\comrevision{FAC (\%)}} &
            {\comrevision{\textbf{89.31}}} &
            {\comrevision{88.44}} \\
            {FID} &
            {\textbf{8.91}} &
            {9.01} \\
            {User study (\%)} &
            {\textbf{78.67}} &
            {21.33} \\
        \bottomrule
        \end{tabular}
        }
        }
        \normalsize
    \normalsize
    
    \vskip 0.05in
    \caption{Proposed DFI feature extractor effects. Column (1) includes the target (real) images, (2) includes DFI reconstructed images, and (3) includes images reconstructed from a \comrevision{$\mathrm{DFI-VGG16}$} using VGG16 as the feature extractor. Experimental setting and metrics were identical to those for Figure~\ref{figure_igan_compare}. }
    \label{figure_dfi_vgg}
\end{figure}

To confirm the discriminator as a good feature extractor, we \comeng{compared} \comrevision{several DFI versions: original DFI (discriminator with $\mathrm{CN}$ network), $\mathrm{DFI-VGG16}$ (pre-trained VGG16 \citep{ref35} using \texttt{Pool5} feature as $\mathrm{CN}$ network input), and $\mathrm{DFI-ResNet}$ (pre-trained residual based network \citep{he2016deep} using features before GAP as $\mathrm{CN}$ network input). Among the pre-trained models, we empirically observe that $\mathrm{DFI-VGG16}$ outperforms $\mathrm{DFI-ResNet}$ (e.g. ResNet34, ResNet50, and ResNet101) in all quantitative metrics, thus we mainly report $\mathrm{DFI-VGG16}$ results. Please note that VGG16 and ResNet are well-known, powerful feature extractors with an excessive number of parameters, and should be much more powerful feature extractors for general purposes.}

Figure~\ref{figure_dfi_vgg} \comeng{shows} several reconstruction examples with quantitative evaluation results (after 40K training iteration steps) using \com{LPIPS,} FID and the user study. \comeng{Surprisingly, the original DFI produces more accurate reconstructions than the $\mathrm{DFI-VGG16}$ in both qualitative and quantitative comparisons. $\mathrm{DFI-VGG16}$ results are sharp and realistic, similar to the proposed DFI alone approach.} However, considering semantic similarity, the original DFI can restore unique attributes, e.g. mustache, race, age, etc., better than the $\mathrm{DFI-VGG16}$. Although \com{LPIPS and} FID scores from the two methods are quite close, the original DFI significantly outperforms $\mathrm{DFI-VGG16}$ in user study results.

Although the pre-trained VGG16 is a powerful feature extractor in general, the deep generalized strong feature extractor might not outperform the shallow but data specific and well-designed feature extractor for inference mapping using the specific training dataset (CelebA). Most importantly, the pre-trained classifier never experiences the GAN training dataset, and hence cannot exploit training data characteristics. If the VGG16 model was finetuned with GAN training data, we would expect it to exhibit more accurate inference mapping. However, that would be beyond the scope of the current paper because VGG16 already requires many more parameters than the proposed DFI approach. Our purpose was to show that DFI was as powerful as VGG16 although requiring significantly less computing resources without additional overheads required for feature extraction. Quantitative comparisons confirm that the original DFI (utilizing discriminator features) performs better than the $\mathrm{DFI-VGG16}$ (utilizing VGG16 features) when the same training iterations are set. Thus, the original DFI is more efficient than the $\mathrm{DFI-VGG16}$ for inference mapping.

\com{
One might consider that discriminator feature $\mathrm{D}^{\mathrm{f}}$ distributions for real and fake images should not overlap because the discriminator objective is to separate fake images from generated and real images. The distributions may not overlap if the discriminator was trained in a stationary environment or the discriminator defeats the generator, i.e., the generator fails. However, the proposed approach simultaneously trains the generator to deceive the discriminator, hence the GAN training is not stationary. Therefore, if the generator is successfully trained, the generated sample distribution will significantly overlap the real sample distribution, i.e., the generator produces realistic samples. Ideally, training is terminated when the discriminator cannot tell the difference between real and fake images, but for practical GANs, the discriminator is not completely deceived. 
}

\com{
Suppose the generator produces highly realistic fake samples, indistinguishable from real samples. Then $\mathrm{D}^{\mathrm{f}}$ for fake samples will significantly overlap with $\mathrm{D}^{\mathrm{f}}$ for real samples. If the generator is not performing well, e.g. under-training, or small network capacity, $\mathrm{D}^{\mathrm{f}}$ for real and fake samples will not overlap because the discriminator defeated the generator. However, in this situation GAN training fails, i.e., none of the inference algorithms can reconstruct the given image. 
}

\com{
To empirically show that $\mathrm{D}^{\mathrm{f}}$ for real and fake images overlap, Figure~\ref{figure_discriminator_feature_pca} projects $\mathrm{D}^{\mathrm{f}}$ on to the two most significant principal component axes using the LSGAN discriminator. The $\mathrm{D}^{\mathrm{f}}$ for real (blue) and fake images (orange) have significant overlap, with the real sample distribution having wider coverage than for the fake samples due to limited diversity, i.e., mode collapse. Therefore, the discriminator offers a meaningful feature extractor for both real and fake images.
}

\subsection{Toward a high quality DFI} \label{subsection_toward_gap}

\com{
To improve inference mapping accuracy, we modified the DFI by selecting the layer for extracting discriminator features $\mathrm{D}^{\mathrm{f}}$; and increasing the connection network capacity. We first introduce a method to improve $\mathrm{D}^{\mathrm{f}}$ by using a middle level discriminator feature, improving DFI accuracy. Then we investigated inference accuracy with respect to connection network capacity, confirming that higher connection network capacity does not degrade DFI accuracy.
}

Since the discriminator feature is extracted from the last layer of the discriminator, it corresponds to a large receptive field. This is advantageous to learn high level information, but incapable of capturing low level details, such as wrinkles, curls, etc. For reconstruction purposes, this choice is clearly disadvantageous to achieve high quality reconstruction. To resolve this limitation, we transfer knowledge from the intermediate feature map discriminator to the connection network.

In particular, we \comeng{calculated} global average pooling (GAP) \citep{ref_GAP} for the intermediate feature map as the compact representation for the intermediate feature map to achieve computational efficiency. We then \comeng{concatenated} GAP outputs extracted from specific layers of the discriminator with the last discriminator feature. We utilized SNGAN architecture \citep{ref43} for the experiments. 

\begin{figure}[t!]
\centering
    \centering
    \includegraphics[width=\columnwidth]{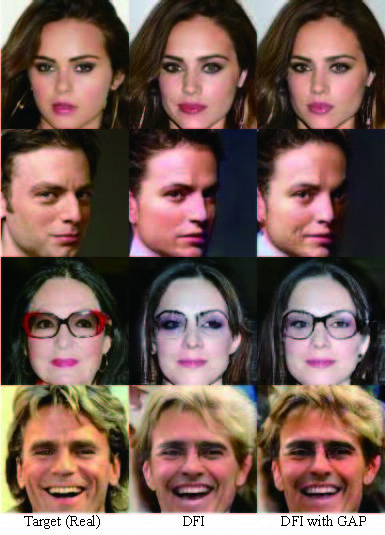} \\
    \caption{Qualitative comparison between DFI and DFI with GAP. Column (1) includes target (real) images, (2) and (3) include images reconstructed by DFI and DFI with GAP on the \texttt{Actv64-1} layer, respectively.}
    \label{figure_dfi_gap_comparison}
\end{figure}

Table ~\ref{figure_sngan_architecture} shows the network architecture and feature map names, Table ~\ref{table_dfi_gap} shows \com{LPIPS and} FID scores for several combinations of extracted GAP layers, and Figure~\ref{figure_dfi_gap_comparison} shows several reconstruction examples for DFI with the GAP layer. Reconstructions from DFI with the GAP layer preserve more attributes attribute, e.g. expressions, eyeglasses, etc.
\com{When utilizing features from a single layer, we found that applying \texttt{Actv64-1} produced the best accuracy in terms of both LPIPS and FID. Combining features from multiple layers, accuracy (LPIPS) increases with increasing number of combinations, whereas FID decreases. Considering fidelity, quality, and computational efficiency, we suggest applying \texttt{Actv64-1} to obtain additional accuracy.}

Although the GAP requires low computational cost, spatial information about the feature is completely missing because GAP reduces the feature map spatial dimension $1 \times 1$. \comeng{Therefore, we should consider} average pooling layer variants, \comeng{considering feature map} spatial information.
To this end, we \comeng{designed an} average pooling to output $(R, R, C)$ feature map, with $R\times R$ final feature map resolution and $C$ is the channel dimension for the intermediate feature map. \comeng{Larger $R$ preserves more feature map spatial information}, and it is equivalent to GAP when $R=1$, i.e., $1 \times 1 \times C$. 
\com{We used the \texttt{Actv64-1} layer in this experiment, since that provided the highest score in single layer combination as well as the FID score.}

Table ~\ref{table_dfi_ap_variation} shows \com{LPIPS and }FID scores corresponding to the average pooling layer using the final $R\times R$ resolution feature map. Thus, average pooling preserving spatial information can empirically improve \com{both fidelity and quality} compared with GAP. However, \com{both scores} increase when $R>4$. We \comeng{suggest this is} due to the large number of parameters, which leads to DFI overfitting the training data.

\begin{table}[t!] 
\small
  \centering
  \resizebox{0.9\columnwidth}{!}{
    \begin{tabular}{c | c c c | c c}
    \toprule
        DFI & \texttt{Actv128-0} & \texttt{Actv64-1}  & \texttt{Actv32-1} & {LPIPS} & {FID} \\
        \midrule
        {\checkmark} &{}           & {}           & {}            & {0.1931} & {8.91} \\
        {\checkmark} &{\checkmark} & {}           & {}            & {0.1920} & {10.34} \\
        {\checkmark} &{}           & {\checkmark} & {}            & {0.1903} & {\textbf{8.54}} \\
        {\checkmark} &{}           & {}           & {\checkmark}  & {0.1917} & {8.83} \\
        {\checkmark} &{\checkmark} & {\checkmark} & {}            & {0.1902} & {9.78} \\
        {\checkmark} &{}           & {\checkmark} & {\checkmark}  & {0.1897} & {9.35} \\
        {\checkmark} &{\checkmark} & {\checkmark} & {\checkmark}  & {\textbf{0.1887}} & {10.32} \\
        \bottomrule
    \end{tabular}
    }
    \vskip 0.1in
    \caption{Reconstruction performance for combinations of extracted layers of GAP (top).
    {\comrevision{LPIPS and}} FID scores are the average of the best performance among 50K training iterations with 10K steps each trial.}
  \label{table_dfi_gap}
  \vskip -0.2in
  \normalsize
\end{table}

\begin{table}[t!] 
\small
  \centering
    \resizebox{0.6\columnwidth}{!}{
    \begin{tabular}{c c | c c}
    \toprule
        {DFI} & {Average pooling} & {LPIPS} & {FID} \\
        \midrule
        {\checkmark} & {1$\times$1} & {0.1903}  & {8.54} \\
        {\checkmark} & {2$\times$2} & {0.1894}  & {8.29} \\
        {\checkmark} & {4$\times$4} & {0.1835}  & {8.26} \\
        {\checkmark} & {8$\times$8} & {0.1858}  & {8.77} \\
        \bottomrule
    \end{tabular}
    }
    \vskip 0.1in
    \caption{Reconstruction performance with respect to average pooled feature map size, $R$. Average pooling was conducted on feature map \texttt{Actv64-1}. 1$\times$1 denotes GAP and R$\times$R denotes average pooling with the feature map. FID scores are average of the best performance among 50K training iterations with 10K steps each trial. Bold values indicate Top-1 scores for each metric.}
  \label{table_dfi_ap_variation}
  \vskip -0.2in
  \normalsize
\end{table}

\begin{table}[t!]
    \small
    \vskip 0.1in
    \begin{minipage}{0.49\textwidth}
        \centering
        \resizebox{0.8\textwidth}{!}{
            \begin{tabular}{l l}
                \toprule
                    \multicolumn{2}{l}{\textbf{Generator}} \\
                    \multicolumn{2}{l}{$z \in \mathbb{R}^{128} \sim \mathcal{N}(0, I)$} \\
                    \multicolumn{2}{l}{Fully connected layer $\rightarrow 16 \times 16 \times 512$ } \\
                    \multicolumn{2}{l}{4$\times$4, stride=2 transposed conv. BN 256 ReLU} \\
                    \multicolumn{2}{l}{4$\times$4, stride=2 transposed conv. BN 128 ReLU} \\
                    \multicolumn{2}{l}{4$\times$4, stride=2 transposed conv. BN \ \ 64 ReLU} \\
                    \multicolumn{2}{l}{3$\times$3, stride=1 conv. 3 Tanh} \\
                    {} & {} \\
                    \multicolumn{2}{l}{\textbf{Discriminator}} \\
                    {} & {RGB images $\in \mathbb{R}^{128 \times 128 \times 3}$} \\
                    \texttt{Actv128-0} & {3$\times$3, stride=1 conv \  64 Leaky ReLU} \\
                    \texttt{Actv64-0} & {4$\times$4, stride=2 conv \ 64 Leaky ReLU} \\
                    \texttt{Actv64-1} & {3$\times$3, stride=1 conv 128 Leaky ReLU} \\
                    \texttt{Actv32-0} & {4$\times$4, stride=2 conv 128 Leaky ReLU} \\
                    \texttt{Actv32-1} & {3$\times$3, stride=1 conv 256 Leaky ReLU} \\
                    \texttt{Actv16-0} & {4$\times$4, stride=2 conv 256 Leaky ReLU} \\
                    {\texttt{Actv16-1} (last feature)} & {3$\times$3, stride=1 conv 512 Leaky ReLU} \\
                    {} & {Fully connected layer $\rightarrow$ 1} \\
                \bottomrule
            \end{tabular}
        }
        \vskip 0.1in
    \end{minipage}
    \vskip 0.1in
    \normalsize
    \caption{Network architecture for SNGAN, using architecture is based on Table 3 of \cite{ref43}.}
    \label{figure_sngan_architecture}
\end{table}

\begin{table}[t!] 
\small
  \centering
    \resizebox{0.6\columnwidth}{!}{
    \begin{tabular}{c | c c}
    \toprule
        {No. FC layers} & {LPIPS} & {FID} \\
        \midrule
        {1} & 0.1942 & {8.96} \\
        {2 (default)} & 0.1931 & {8.91} \\
        {3} & 0.1943 & {8.92} \\
        {4} & 0.1948 & {8.85} \\
        {5} & 0.1946 & {8.73} \\
        \bottomrule
    \end{tabular}
    }
    \vskip 0.1in
    \caption{\com{Reconstruction performance with respect to number of FC layers in the connection network. LPIPS and FID scores are average score of best performance among 50K training iterations with 10K steps each trial.}}
  \label{table_dfi_lin_variation}
  \vskip -0.2in
  \normalsize
\end{table}

\com{
The DFI modeling power solely depends on the connection network capacity because both the generator and discriminator are fixed when training the connection network. Training high capacity networks commonly suffer from overfitting with limited datasets. \comeng{Therefore, the proposed} inference algorithm may also experience overfitting on training data if the high capacity model \comeng{was} selected for the connection network. Fortunately, in the training scenario using the proposed latent reconstruction loss, we can utilize unlimited training samples because their seed, i.e., a latent code, can be drawn from a continuous prior distribution and their images can be created by the generator. \comeng{Thus, regardless of network capacity}, we \comeng{will} have sufficient training data to avoid overfitting. \comeng{Consequently}, the network capacity \comeng{(provided it includes more than two FC layers)} does not affect inference mapping accuracy.}

\com{
To verify this, we \comeng{investigated} inference accuracy with respect to the number of connection network layers, i.e., connection network capacity. The default setting for other experiments \comeng{reported here was} two FC layers. Table~\ref{table_dfi_lin_variation} summarizes LPIPS and FID scores \comeng{for various numbers of} FC layers in the connection network. Thus we experimentally verify that connection network complexity does not significantly influence inference accuracy. 
}

\subsection{SCGAN experimental results}

\begin{figure*}[t!]
\centering
    \centering
    \includegraphics[width=\textwidth]{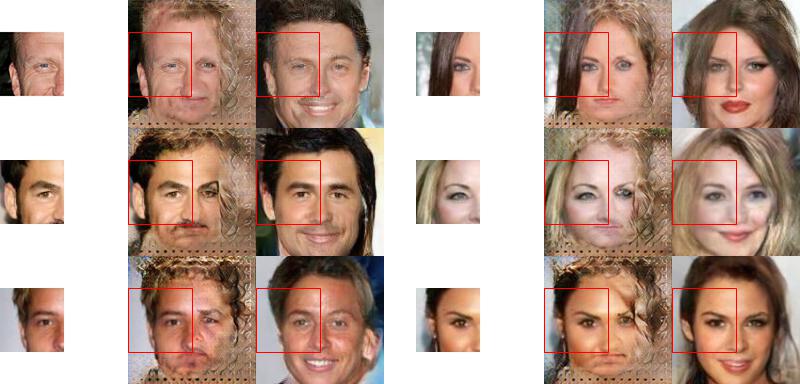} \\
    \caption{\com{Proposed SCGAN compared with PICNet state-of-the-art image completion technique. Column (1) and (4) include input images, (2), (3), (5), and (6) include PICNet and SCGAN generations, respectively.}}
    \label{figure_comparison_image_completion}
\end{figure*}

We \comeng{verified} spatially conditioned image generation feasibility using the proposed SCGAN approach for the CelebA~\citep{ref14} and cat head~\citep{zhang2008cat} dataset. All experiments set center image size (input) = $(64, 64, 3)$ and full image size = $(128, 128, 3)$. We assigned the input patch location to the middle left for the CelebA dataset and top left for the cat head dataset. Latent vector dimension = 128 for ${z}_{full}$ and 64 for both ${z}_{center}$ and ${z}_{edge}$. SCGAN baseline architecture \comeng{was} built upon SNGAN \citep{ref43}, where only spectral normalization was applied to the discriminator. Throughout all SCGAN experiments, we used hyperparameter $\alpha=10$ for ${L}^{recon}$ and $\gamma=10$ for ${L}^{GP}$.

Two evaluation criteria \comeng{were employed} for spatially conditioned image generation: reconstruction accuracy and generation quality. To assess reconstruction quality \comeng{we adopted} LPIPS and FID.
First, we measured LPIPS and FID scores between ${x}_{center}$ and ${y}_{center}$, reconstructed by the proposed DFI inference algorithm, using 10k test images from CelebA and 1k test images from the cat head dataset. These scores, $(0.1673, 31.24)$ and $(0.1669, 32.64)$, respectively, \comeng{served} as the baseline for SCGAN reconstruction quality. We then \comeng{calculated} both scores between ${x}_{center}$ and ${y}_{crop}$ (reconstructed by SCGAN), achieving $(0.1646, 31.70)$ and $(0.1653, 33.03)$ respectively, which are comparable with the baseline LPIPS and FID scores. Hence SCGAN reconstruction ability is similar to the proposed inference algorithm.

To qualitatively assess generation quality, we examined whether generated images \comeng{were} diverse, semantically consistent with the reconstructed image, and visually pleasing. Figures~\ref{figure_scgan_celeba} and \ref{figure_scgan_cat} show example spatially conditioned images using SCGAN. Row (1) includes input images (inside box) with their surrounding regions, and rows (2) and (3) include various image generation results from the same input, i.e., the same input latent vector, $\hat{z}_{center}$, but with a different ${z}_{edge}$ latent vectors. 
Figure~\ref{figure_scgan_celeba} shows six generated results for different ${z}_{edge}$ are clearly different from each other, presenting various facial shapes, hairstyles, or lips for the same input. However, all reconstructions have acceptable visual quality and match input image semantics well in terms hair color, skin tone, or eye and eyebrow shape. 
Figure~\ref{figure_scgan_cat} shows four generated cat head dataset reconstructions with similar tendencies to CelebA results. Each cat has a different face shape, hair color, and expression, with reasonable visual quality. However, the input is correctly reconstructed, and the generated surroundings are semantically seamless with the input. 

Thus, SCGAN successfully controlled spatial conditions by assigning input position, producing various high quality images.

\com{
Finally, we compared the proposed approach with the PICNet state-of-the-art image completion technique~\citep{zheng2019pluralistic} under the same conditions, as shown in Figure~\ref{figure_comparison_image_completion}. SCGAN can generate realistic entire faces, whereas PICNet cannot maintain consistent quality across the entire image region. This is due to the surrounding regions requiring extrapolation, whereas PICNet image completion is designed to solve image interpolation. \comeng{Unlike various image completion models such as PICNet, SCGAN possesses the strong generation capability of GANs, producing the images from the latent codes, despite it can faithfully keep the input patch by utilizing inference mapping. As a result, SCGAN solves image extrapolation, which is not possible by previous image completion models.}
}
\section{Conclusion}
This study \comeng{proposed} an acyclic inference algorithm to improve inference accuracy with minimal training overhead. We \comeng{introduced} discriminator feature based inference (DFI) to map discriminator features to the latent vectors. Extensive experimental evaluations \comeng{demonstrated} that the proposed DFI approach outperforms current methods, accomplishing semantically accurate and computationally efficient inference mapping.

We believe the accuracy gain is achieved by the well-defined objective function, i.e., latent reconstruction loss; and the powerful feature representation from the discriminator. The computational problem \comeng{was} simplified into deriving the mapping from low dimensional representation to another low dimensional representation by adopting discriminator features. Consequently, the proposed approach also provides computational efficiency in training by significantly reducing training parameters.

We also \comeng{introduced} a novel conditional image generation algorithm (SCGAN), incorporating the proposed DFI approach. SCGAN can generate spatially conditioned images using accurate semantic information inferred from the proposed inference mapping. We experimentally \comeng{demonstrated} that spatial information about the image can be used as a conditional prior, in contrast to traditional priors, e.g., class labels or text. We expect the proposed model architecture can be extended to solve image extrapolation and editing problems.

% \section*{Acknowledgment}
% This research was supported by the Basic Science Research Program through the National Research Foundation of Korea funded by the Korean Government (grant NRF-2019R1A2C2006123), the MSIT (Ministry of Science and ICT), Korea, under the “ICT Consilience Creative Program” (IITP-2019-2017-0-01015) supervised by the IITP (Institute for Information \& communications Technology  Planning \& Evaluation), and also by ICT R\&D program of MSIP/IITP. [ R7124-16-0004, Development of Intelligent Interaction Technology Based on Context Awareness and Human Intention Understanding ]. 

\bibliographystyle{spbasic}      % basic style, author-year citations
\bibliography{reference}   % name your BibTeX data base

\end{document}